\documentclass[lettersize,journal]{IEEEtran}
\usepackage{amsmath,amsfonts}
\usepackage{algorithmic}
\usepackage{algorithm}
\usepackage{array}
\usepackage[caption=false,font=normalsize,labelfont=sf,textfont=sf]{subfig}
\usepackage{textcomp}
\usepackage{stfloats}
\usepackage{url}
\usepackage{verbatim}
\usepackage{graphicx}
\usepackage{relsize}
\usepackage{colortbl}
\usepackage{cite}
\usepackage{dsfont}
\usepackage{hyperref}
\newcommand{\ie}{\textit{i.e.}}
\newcommand{\eg}{\textit{e.g.}}
\newcommand{\etal}{\textit{et al.}}
\newcommand{\midoplus}{\mathlarger{\mathlarger{\oplus}}}
\usepackage{graphicx}
\usepackage{amsmath}
\usepackage{balance}
\usepackage{amssymb}
\usepackage{booktabs}
\usepackage{bbm}
\usepackage{bm}
\usepackage[normalem]{ulem}
\usepackage{amssymb}
\usepackage[none]{hyphenat}
\usepackage{pifont}
\usepackage{marvosym}
\usepackage{xcolor}
\usepackage{multirow}
\definecolor{fit_curve_green}{RGB}{169,209,142}
\definecolor{supple_pink}{RGB}{245,151,155}
\definecolor{supple_blue}{RGB}{0,32,96}
\definecolor{supple_red}{RGB}{192,0,0}
\begin{document}

\title{HAC++: Towards 100X Compression of 3D Gaussian Splatting}

\author{Yihang Chen, Qianyi Wu\textsuperscript{\textdagger}, Weiyao Lin\textsuperscript{\textdagger}, Mehrtash Harandi, Jianfei Cai,~\IEEEmembership{Fellow,~IEEE}
        % <-this % stops a space

\thanks{Yihang Chen, Weiyao Lin are with Shanghai Jiao Tong University. Email: \{yhchen.ee, wylin\}@sjtu.edu.cn}
\thanks{Yihang Chen, Qianyi Wu, Mehrtash Harandi, Jianfei Cai are with Monash University. Email: \{yihang.chen, qianyi.wu, mehrtash.harandi, jianfei.cai\}@monash.edu}
\thanks{\textsuperscript{\textdagger}Corresponding authors: Qianyi Wu and Weiyao Lin.}
}

% \markboth{IEEE TPAMI SUBMISSION}%
% {Shell \MakeLowercase{\textit{et al.}}: HAC++}

\markboth{Journal of \LaTeX\ Class Files,~Vol.~14, No.~8, August~2021}%
{Shell \MakeLowercase{\textit{et al.}}: A Sample Article Using IEEEtran.cls for IEEE Journals}


\maketitle

\begin{abstract}
  3D Gaussian Splatting (3DGS) has emerged as a promising framework for novel view synthesis, boasting rapid rendering speed with high fidelity. However, the substantial Gaussians and their associated attributes necessitate effective compression techniques. Nevertheless, the sparse and unorganized nature of the point cloud of Gaussians (or anchors in our paper) presents challenges for compression. To achieve a compact size, we propose HAC++, which leverages the relationships between unorganized anchors and a structured hash grid, utilizing their mutual information for context modeling. Additionally, HAC++ captures intra-anchor contextual relationships to further enhance compression performance. 
  To facilitate entropy coding, we utilize Gaussian distributions to precisely estimate the probability of each quantized attribute, where an adaptive quantization module is proposed to enable high-precision quantization of these attributes for improved fidelity restoration. Moreover, we incorporate an adaptive masking strategy to eliminate invalid Gaussians and anchors.
  Overall, HAC++ achieves a remarkable size reduction of over $100\times$ compared to vanilla 3DGS when averaged on all datasets, while simultaneously improving fidelity. It also delivers more than $20\times$ size reduction compared to Scaffold-GS. Our code is available at https://github.com/YihangChen-ee/HAC-plus.
\end{abstract}


\section{Introduction}

\IEEEPARstart{O}{ver} the past few years, significant advancements have been made in novel view synthesis. Neural Radiance Field (NeRF)~\cite{NeRF} introduced a groundbreaking approach for establishing 3D representations from input images by rendering colors through the accumulation of RGB values along sampling rays using an implicit Multilayer Perceptron (MLP). However, the large volume of the MLP and the extensive ray point sampling have become bottlenecks, hindering both training and rendering speeds. Advancements in NeRF~\cite{INGP, TensoRF, K-planes} introduce feature grids to enhance the rendering process by involving learnable explicit representations, facilitating faster rendering speeds using swifter MLPs. Nevertheless, these methods still suffer from relatively slow rendering due to the dense ray point sampling.

In this context, a new type of 3D representation, 3D Gaussian Splatting (3DGS)\cite{3DGS}, has recently emerged. 3DGS describes 3D scenes using learnable explicit Gaussians. These Gaussians are derived from Structure-from-Motion (SfM)~\cite{SfM}, which reconstructs 3D structures from images and views. Enhanced with learnable shape and appearance parameters, the Gaussians can be efficiently splatted onto 2D planes using tile-based rasterization~\cite{raster}, enabling rapid and differentiable rendering. By eliminating the computational burden of the dense point sampling, 3DGS achieves a significant boost in rendering speed. Its advantages of fast rendering and high photo-realistic fidelity have driven its rapid adoption.

Despite these advancements, 3DGS faces the challenge of massive Gaussian primitives (\eg, millions of Gaussians for city-scale scenes), resulting in large storage requirements (\eg, a few GigaBytes (GB) to store the Gaussian attributes for each scene~\cite{BungeeNeRF, mip360}). This issue motivates the exploration of effective compression techniques for 3DGS.

\begin{figure*}[t]
    \centering
    \includegraphics[width=0.95\linewidth]{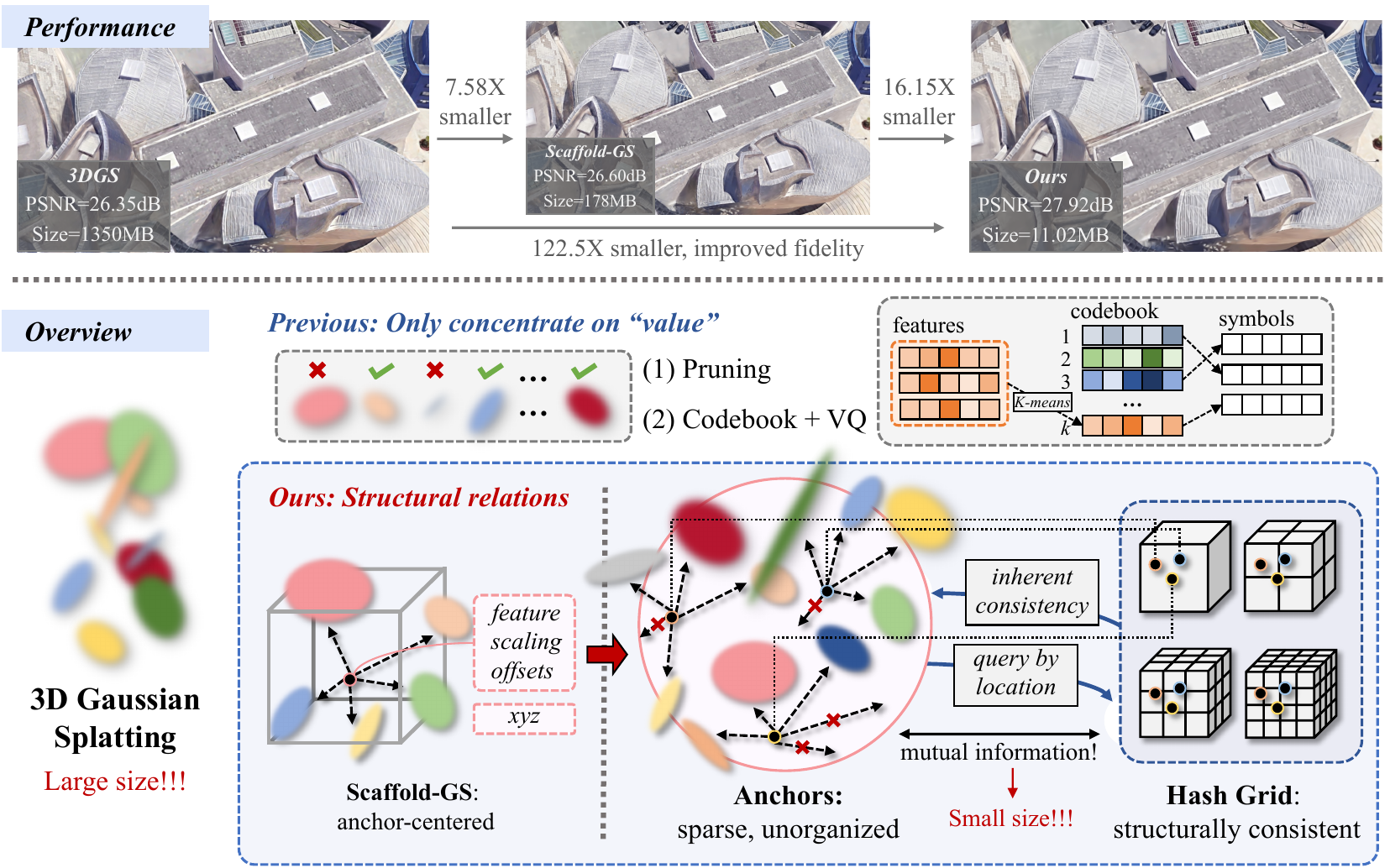}
    \caption{\textbf{Top}: A toy example showcasing the effectiveness of our method, which reduces the size of the vanilla 3D Gaussian Splatting (3DGS) model by $122.5\times$ (or $16.15\times$ compared to the SoTA Scaffold-GS~\cite{scaffold}), with similar or better fidelity. \textbf{Bottom}: Most existing 3DGS compression methods typically focus on parameter ``values'' through pruning or vector quantization, overlooking structural relations among Gaussians. Scaffold-GS~\cite{scaffold} introduces anchors to cluster Gaussians and neural-predict their attributes but treats each anchor independently. In contrast, our method leverages the inherent consistencies among anchors via a structured hash grid, enabling a significantly more compressed 3DGS representation.}
    \label{fig:teaser}
\end{figure*}

However, compressing 3D Gaussians is inherently challenging~\cite{chen2024survey, fei2024survey, 3dgszip} due to their sparse and unorganized nature. Various methods have been proposed to address this issue~\cite{Lee, Simon, EAGLES, Lightgaussian} by utilizing Gaussian pruning or codebook-based vector quantization, as illustrated in Fig.~\ref{fig:teaser}. However, they share a common limitation: they focus on parameter ``values'' while neglecting the redundancies inherent in structural relations.
Structural relations play a crucial role in uncovering redundancies among parameters, which is a principle that has been effectively demonstrated in image compression~\cite{cheng2020learned, he2021checkerboard}. While such relations are relatively easier to identify for images, leveraging them for 3D Gaussians remains a challenge. To address this, Scaffold-GS~\cite{scaffold} introduces anchors to cluster related 3D Gaussians and uses neural predictions to infer their attributes from attributes of the anchor, achieving substantial storage savings. Despite advancements, Scaffold-GS treats each anchor independently, leaving anchors sparse, unorganized, and difficult to compress due to their point-cloud-like nature.

We draw inspiration from the NeRF series\cite{NeRF}, which leverage well-organized feature grids~\cite{INGP, TensoRF} to represent 3D space. We pose the question: \emph{Are there inherent relations between the attributes of unorganized anchors in Scaffold-GS and the structured feature grids?} Our answer is affirmative since we observe large mutual information between anchor attributes and the hash grid features. Based on this insight, we propose our HAC++ method. The core idea is a Hash-grid Assisted Context (HAC) framework to jointly learn a structured and compact hash grid (with binarized hash parameters), and utilize it for context modeling of anchor attributes. Specifically, with Scaffold-GS~\cite{scaffold} as our base model, for each anchor, we query the hash grid by the anchor location to obtain an interpolated hash feature, which is then used to predict the value distributions of anchor attributes. 

Beyond HAC modeling, we integrate an intra-anchor context model to effectively capture the internal contextual relationships of anchors, which provides auxiliary information that further enhances the context accuracy of HAC. It leverages channel-wise redundancies in the anchor feature by employing a chunk-by-chunk prediction flow under a causal process. This intra-anchor context is combined with HAC using a Gaussian Mixture Model (GMM) to provide auxiliary information, which facilitates entropy coding such as Arithmetic Coding (AE)~\cite{AE} for a highly compressed representation. 
Note that we opt for Scaffold-GS as our base model due to its excellent characteristics of high fidelity and efficient speed, particularly in real-world large-scale datasets. The compression performance also benefits from its feature-based design to achieve a much smaller size.

Additionally, we introduce an Adaptive Quantization Module (AQM) that dynamically adjusts quantization step sizes for different anchor attributes, achieving a balance between retaining original information and reducing entropy. Learnable masks are employed to exclude invalid Gaussians and anchors, further improving the compression ratio. Specifically, HAC++ incorporates the mask information directly into the rate calculation in a differentiable manner, eliminating the need for an extra loss term that may otherwise impact the optimal RD trade-off. Furthermore, anchor-level masks are explicitly deduced from offset masks to proactively reduce the number of anchors. This approach adaptively determines the optimal mask ratios across different bit rate points.

This paper is an extension of our \textit{ECCV}'2024 work, HAC~\cite{HAC}. By revisiting the original design and incorporating more advanced techniques, we have significantly enhanced its performance. 
Moreover, this paper offers a comprehensive analysis of the proposed approach. We believe this work can push the performance of 3DGS compression a step forward.

Our main contributions can be summarized as follows:
\begin{enumerate}
    \item We propose HAC++, an innovative approach that bridges the relationship between the structured hash grid and unorganized 3D Gaussians (or anchors in Scaffold-GS), achieving effective context modeling for compression.
    
    \item To facilitate efficient entropy encoding of anchor attributes, we propose to use the interpolated hash feature to neural-predict the value distribution of anchor attributes as well as neural-predicting quantization step refinement with AQM. The intra-anchor context model is further utilized to exploit internal information of anchors to improve the context accuracy. We also incorporate adaptive offset masks to prune out ineffective Gaussians and anchors, further slimming the model. 
    
    \item Extensive experiments on five datasets demonstrate the effectiveness of HAC++ and its individual technical components. Our method achieves an average compression ratio of over $100\times$ compared to the vanilla 3DGS model when averaged over all datasets and $20\times$ compared to the base model Scaffold-GS, while maintaining or even improving fidelity. Comprehensive analyses are conducted to provide in-depth technical insights.
    
\end{enumerate}

\section{Related Work}

\noindent\textbf{Neural Radiance Field and Its Compression.} Neural Radiance Field (NeRF)\cite{NeRF} has promoted the novel view synthesis by using learnable implicit MLPs to render 3D scenes via $\alpha$-composed accumulation of RGB values along a ray. However, the dense sampling of query points and reliance on large MLPs hinder the real-time rendering ability. To address this issue, methods such as Instant-NGP~\cite{INGP}, TensoRF~\cite{TensoRF}, K-planes~\cite{K-planes}, and DVGO~\cite{DVGO} adopt explicit grid-based representations, reducing the size of the MLPs to improve the training and rendering efficiency, which, however, come at the cost of increased storage requirements.

To alleviate storage demands, compression techniques for explicit representations have been extensively developed, which fall into two main categories: \textit{value-based} and \textit{structural-relation-based}.
Value-based methods, such as pruning~\cite{VQRF, Re:NeRF} and codebooks~\cite{VQRF, CompactNeRF}, aim to reduce the number of parameters and streamline the model. Pruning removes insignificant parameters based on threshold comparisons, while codebooks cluster parameters with similar values into shared codewords for more efficient storage. Additionally, quantization~\cite{BiRF} and entropy constraints~\cite{SHACIRA} reduce the bit-length for parameter representation by minimizing their information content. However, these approaches treat parameters independently, overlooking the mutual relationships that could further enhance compression efficiency.
In contrast, structural-relation-based methods leverage spatial redundancies in well-structured grids to improve the compression performance. Techniques such as wavelet decomposition~\cite{MaskDWT}, rank-residual decomposition~\cite{CCNeRF}, and spatial prediction~\cite{SPC-NeRF} have shown great promise. Built on Instant-NGP~\cite{INGP}, CNC~\cite{cnc2024} demonstrates the potential of leveraging structural information to achieve substantial RD performance gains. Similarly, NeRFCodec~\cite{NeRFCodec} represents neural fields using regular planes and vectors, applying entropy constraints for compression. CodecNeRF~\cite{codecnerf} employs a feed-forward approach to directly compress the neural radiance fields.

\noindent\textbf{3D Gaussian Splatting and its compression.} 3DGS~\cite{3DGS} has innovatively addressed the challenge of slow training and rendering in NeRF while maintaining high-fidelity quality by representing 3D scenes with explicit 3D Gaussians endowed with learnable shape and appearance attributes. By adopting differentiable splatting and tile-based rasterization~\cite{raster}, 3D Gaussians are optimized during training to best fit their local 3D regions. Despite its advantages, the substantial Gaussians and their associated attributes necessitate effective compression techniques.

Unlike the well-structured feature grids in NeRF-based representations, Gaussians in 3DGS are inherently sparse and unorganized, posing challenges in establishing structural relationships~\cite{3dgszip, wu2024recent}. Consequently, most compression methods follow the NeRF compression pipeline and primarily focus on reducing the number of parameters through \textit{value-based} techniques. As redundancies exist among Gaussians, pruning approaches have been extensively explored~\cite{Lee, Lightgaussian, Navaneet, Reduced3DGS, trimming}, by employing strategies like trainable masks, gradient-informed thresholds, view-dependent metrics, and other importance-evaluation mechanisms. Beyond the entire Gaussians, Gaussian attributes such as Sphere Harmonics (SH) coefficients can also be partially adjusted or pruned, as demonstrated in~\cite{SOG, Reduced3DGS}. Additionally, codebooks~\cite{Lee, Simon, Lightgaussian, Navaneet, RDOGaussian} and entropy constraints~\cite{EAGLES} have also been widely utilized for parameter reduction and compression. On the other hand,
\textit{structural-relation-based} techniques have gained much attention. \cite{SOG} employs dimensional collapsing to organize Gaussians into a structured 2D grid for compression. Combining pruning with structural relations, SUNDAE~\cite{SUNDAE} leverages spectral graph modeling, while Mini-Splatting~\cite{mini} utilizes spatial relation-aware spawn techniques to achieve compact representations. IGS~\cite{IGS} predicts attributes of unstructured 3D locations using a multi-level grid for a compact 3D modeling.
Notably, Scaffold-GS~\cite{scaffold} introduces an anchor-centered framework that leverages features for neural prediction of Gaussian attributes, achieving parameter reduction with improved fidelity. Built on Scaffold-GS, HAC~\cite{HAC} explores inherent spatial redundancies among anchors using compact binarized hash grids for parameter quantization and entropy modeling. Concurrently, ContextGS~\cite{Contextgs} and CompGS~\cite{liu2024compgs} both utilize context-aware designs by modeling hierarchical anchor relations or anchor-Gaussian relations. 

While existing methods effectively establish relationships among anchors, they often overlook the internal redundancies within anchors. Additionally, the unique anchor-based structure of Scaffold-GS presents opportunities for designing optimized pruning strategies.
To address these limitations, we adopt Scaffold-GS as our base model, leveraging both inter- and intra-anchor contexts combined with an improved pruning strategy to enable a more compact and efficient 3DGS representation.

\section{Methods}
\begin{figure*}[t]
    \centering
    \includegraphics[width=1.0\linewidth]{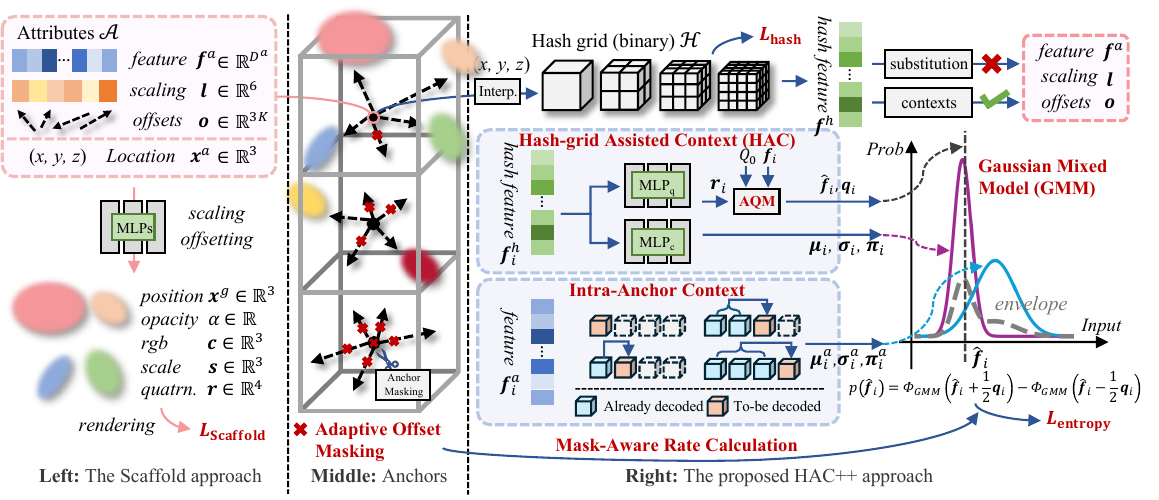}
    \caption{\textbf{Overview of our HAC++ framework.} Built upon Scaffold-GS~\cite{scaffold} (\textbf{left}), which introduces anchors and their attributes to neural-predict 3D Gaussian attributes, HAC++ enhances compression performance by modeling both inter- and intra-anchor relations. \textbf{Right}: HAC++ consists of a Hash-grid Assisted Context (HAC) and an Intra-Anchor Context. HAC learns a structured and compact hash grid (binarized for each parameter), which is queried at anchor locations to generate interpolated hash features $\bm{f}^h$. Instead of directly replacing anchor features, $\bm{f}^h$ serves as context for predicting the value distributions of anchor attributes, which is crucial for entropy coding. The intra-anchor context further improves prediction accuracy by eliminating internal redundancies of anchors. Additionally, HAC outputs ${\bm{r}}$ for the Adaptive Quantization Module (AQM), which quantizes anchor attribute values into a finite set for entropy coding. An Adaptive Offset Masking strategy (\textbf{middle}) is integrated to prune redundant Gaussians and anchors, enhancing the model’s pruning efficiency across different rate points.}
    \label{fig:main_method}
\end{figure*}

In Fig.~\ref{fig:main_method}, we conceptualize our HAC++ framework. In particular, HAC++ is built based on Scaffold-GS~\cite{scaffold} (Fig.~\ref{fig:main_method}~{\textbf{left}}), which introduces anchors with their attributes $\mathcal{A}$ (feature, scaling and offsets) to cluster and neural-predict 3D Gaussian attributes (opacity, RGB, scale, and quaternion). At the core of our HAC++, it consists of a Hash-grid Assisted Context (HAC) (Fig.~\ref{fig:main_method}~{\textbf{right}}) for modeling of inter-anchor relations and an intra-anchor context model (Fig.~\ref{fig:main_method}~{\textbf{right}}) that exploits internal redundancies of anchors. 
Specifically:
\textbf{(1) Hash-grid Assisted Context (HAC):}
HAC jointly learns a structured compact hash grid (binarized for each parameter), which can be queried at any anchor location to obtain an interpolated hash feature $\bm{f}^h$ (Fig.~\ref{fig:main_method}~\textbf{right}). Instead of directly replacing the anchor features, $\bm{f}^h$ serves as context to predict the value distributions of anchor attributes, which is essential for subsequent entropy coding via Arithmetic Coding (AE). Additionally, HAC takes $\bm{f}^h$ as input and outputs ${\bm{r}}$ for the adaptive quantization module (AQM) (quantizes anchor attribute values into a finite set) and the Gaussian parameters ($\bm{\mu}^\text{s}$ and $\bm{\sigma}^\text{s}$) for modeling the value distributions of anchor attributes, from which we can compute the probability of each quantized attribute value for AE.
\textbf{(2) Intra-Anchor Context:}
The intra-anchor context eliminates internal redundancies of anchors, serving as auxiliary information to enhance the prediction accuracy of HAC's value distributions, thereby improving overall compression performance.
\textbf{(3) Adaptive Offset Masking:}
This module is utilized to prune redundant Gaussians and anchors. Specifically, the adaptive offset masking is directly integrated into the rate calculation in a differentiable manner (Fig.~\ref{fig:main_method}~\textbf{middle}), removing the need for an additional regularization loss term. This design enables the framework to achieve an optimal masking ratio for different rate points.
In the following sections, we provide a background overview and then detail each technical component of our HAC++ framework.

\subsection{Preliminaries}

\noindent\textbf{3D Gaussian Splatting (3DGS)}~\cite{3DGS} represents a 3D scene using numerous Gaussians and renders viewpoints through a differentiable splatting and tile-based rasterization. Each Gaussian is initialized from SfM and defined by a 3D covariance matrix $\bm{\Sigma}\in\mathbb{R}^{3\times3}$ and location (mean) $\bm{\mu}\in\mathbb{R}^{3}$,

\begin{equation}
    G(\bm{x}) = \exp{\left(-\frac{1}{2}(\bm{x}-\bm{\mu})^\top\bm{\Sigma}^{-1}(\bm{x}-\bm{\mu})\right)}\;,
\end{equation}
where $\bm{x}\in\mathbb{R}^{3}$ is a random 3D point, and $\bm{\Sigma}$ is defined by a diagonal matrix $\bm{S}\in\mathbb{R}^{3\times3}$ representing scaling
and rotation matrix $\bm{R}\in\mathbb{R}^{3\times3}$ to guarantee its positive semi-definite characteristics, such that $\bm{\Sigma}=\bm{R}\bm{S}\bm{S}^\top\bm{R}^\top$.
To render an image from a random viewpoint, 3D Gaussians are first splatted to 2D, and render the pixel value $\bm{C}\in\mathbb{R}^{3}$ using $\alpha$-composed blending,

\begin{equation}
    \bm{C} = \sum_{i\in I} {\bm{c}_i\alpha_i\prod_{j=1}^{i-1}\left(1-\alpha_j\right)}
\end{equation}
where $\alpha\in\mathbb{R}$
measures the opacity of each Gaussian after 2D projection, $\bm{c}\in\mathbb{R}^{3}$ is view-dependent color modeled by Spherical Harmonic (SH) coefficients, and $I$ is the number of sorted Gaussians contributing to the rendering.

\noindent\textbf{Scaffold-GS}~\cite{scaffold} adheres to the framework of 3DGS and introduces a more storage-friendly and fidelity-satisfying anchor-based approach. It utilizes anchors to cluster Gaussians and deduce their attributes from the attributes of attached anchors through MLPs, rather than directly storing them. Specifically, each anchor consists of a location $\bm{x}^a\in\mathbb{R}^3$ and anchor attributes $\mathcal{A}=\{\bm{f}^a\in\mathbb{R}^{D^a}, \bm{l}\in\mathbb{R}^6, \bm{o}\in\mathbb{R}^{3K}\}$, where each component represents anchor feature, scaling and offsets, respectively. During rendering, $\bm{f}^a$ is inputted into MLPs to generate attributes for Gaussians, whose locations are determined by adding $\bm{x}^a$ and $\bm{o}$, where $\bm{l}$ is utilized to regularize both locations and shapes of the Gaussians. While Scaffold-GS has demonstrated effectiveness via this anchor-centered design, we contend there is still significant redundancy among inherent consistencies of anchors that we can fully exploit for a more compact 3DGS representation.

\subsection{Bridging Anchors and Hash Grid}
We begin the analysis by intuitively considering neighboring Gaussians share similar parameters inferred from anchor attributes. This initial perception leads us to assume anchor attributes are also consistent in space. Our main idea is to leverage the well-structured hash grid to unveil the inherent spatial consistencies of the unorganized anchors. Please also refer to experiments in~\ref{fig:bit_allocation} to observe this consistency.
To verify mutual information between the hash grid and anchors, we first explore substituting 
anchor features $\bm{f}^a$ with hash features $\bm{f}^h$ that are acquired by interpolation using the anchor location $\bm{x}^a$ on the hash grid $\mathcal{H}$,
defined as $\bm{f}^h:= {\rm Interp}(\bm{x}^a, \mathcal{H})$. Here, $\mathcal{H}=\{\bm{\theta}_i^l\in\mathbb{R}^{D^h}|i=1,\dots,T^l|l=1,\dots,L\}$ represents the hash gird, where $D^h$ is the dimension of vector $\bm{\theta}_i^l$, $T^l$ is the table size of the grid for level $l$, and $L$ is the number of levels.
We conduct a preliminary experiment on the Synthetic-NeRF dataset~\cite{NeRF} to assess its performance, as shown in Tab.~\ref{tab:substitution_Synthetic_NeRF}. Direct substitution using hash features appears to yield inferior fidelity and introduces drawbacks such as unstable training (due to its impact on anchor spawning processes) and decreased testing FPS (owing to the extra interpolation operation). These results may further degrade if $\bm{l}$ and $\bm{o}$ are also substituted for a more compact model. Nonetheless, we find the fidelity degradation remains moderate, suggesting the existence of rich mutual information between $\bm{f}^h$ and $\bm{f}^a$. This prompts us to ask: \emph{Can we exploit such mutual relation and use the compact hash features to model the context of anchor attributes $\mathcal{A}$?}
This leads to the context modeling as a conditional probability:
\begin{equation}
\label{eq:prior_probability}
    p(\mathcal{A}, \bm{x}^a, \mathcal{H}) = p(\mathcal{A}|\bm{x}^a, \mathcal{H})\times p(\bm{x}^a, \mathcal{H}) \sim p(\mathcal{A}|\bm{f}^h)\times p(\mathcal{H})
\end{equation}
where $\bm{x}^a$ is omitted in the last term as we assume the independence of $\bm{x}^a$ and $\mathcal{H}$ (it can be anywhere), making $p(\mathcal{H}|\bm{x}^a) \sim p(\mathcal{H})$, and do not employ entropy constraints to $\bm{x}^a$.
According to information theory~\cite{Elements_of_information_theory}, a higher probability corresponds to lower uncertainty (entropy) and fewer bits consumption. Thus, the large mutual information between $\mathcal{A}$ and $\bm{f}^h$ ensures a large $p(\mathcal{A}|\bm{f}^h)$. Our goal is to devise a solution to effectively leverage this relationship. Furthermore, $p(\mathcal{H})$ signifies that the size of the hash grid itself should also be compressed, which can be done by adopting the existing solution for Instant-NGP compression~\cite{cnc2024}.

We underscore the significance of this conditional probability based approach since it ensures both rendering speed and fidelity upper-bound unaffected as it only utilizes hash features to estimate the entropy of anchor attributes for entropy coding but does not modify the original Scaffold-GS structure. In the following subsections, we delve into the technical details of our HAC++ approach, which consists of Hach-grid Assisted Context (HAC) and an intra-anchor context that captures inter- and intra- relation of anchors, respectively.

\begin{table}[t]
    \centering
    \caption{Experimental Results of Directly Substituting Anchor Feature $\bm{f}^a$ with Hash Feature $\bm{f}^h$ on the Synthetic-NeRF Dataset~\cite{NeRF}.}
    \begin{tabular}{cccc}
        \toprule
        {\bf Synthetic-NeRF~\cite{NeRF}} & PSNR$\uparrow$ & SSIM$\uparrow$ & LPIPS$\downarrow$\\
        \midrule
        3DGS~\cite{3DGS} & 33.80 & 0.970 & 0.031\\
        Scaffold-GS~\cite{scaffold} & 33.41 & 0.966 & 0.035\\
        Substituting $\bm{f}^a$ with $\bm{f}^h$ & 32.85 & 0.963 & 0.041\\
        \bottomrule
    \end{tabular}
    \label{tab:substitution_Synthetic_NeRF}
\end{table}

\subsection{HAC: Hash-Grid Assisted Context Framework}
The pipeline of HAC is shown in the right of Fig.~\ref{fig:main_method}, which aim at
minimizing the entropy of anchor attributes $\mathcal{A}$ by eliminating redundancies among anchors with the assistance of hash feature $\bm{f}^h$ (\ie, maximize $p(\mathcal{A}|\bm{f}^h)$), thereby facilitating bit reduction when encoding anchor attributes using entropy coding like AE~\cite{AE}.
As shown in Fig.~\ref{fig:main_method}, anchor locations $\bm{x}^a$ are firstly inputted into the hash grid for interpolation, the obtained $\bm{f}^h$ are then employed as context for $\mathcal{A}$.

\noindent\textbf{Adaptive Quantization Module (AQM)}. 
To facilitate entropy coding, values of $\mathcal{A}$ must be quantized to a finite set. Our experimental studies reveal that binarization, as that in BiRF~\cite{BiRF}, is unsuitable for $\mathcal{A}$ as it fails to preserve sufficient information, with a PSNR of only $31.27$ dB in the Synthetic-NeRF dataset~\cite{NeRF} if we simply binarize all $\bm{f}^a$. Thus, we opt for rounding them to maintain their comprehensive features. To ensure backpropagation, we utilize the ``adding noise'' operation during training and ``rounding'' during testing, as described in \cite{balle2018variational}.

Nevertheless, the conventional rounding is essentially a quantization with a step size of ``1'', which is inappropriate for the scaling $\bm{l}$ and the offsets $\bm{o}$, since they are usually decimal values. To address this, we further introduce an  Adaptive Quantization Module (AQM), which adaptively determines quantization steps. 
In particular, for the $i$th anchor $\bm{x}^a_i$, we denote $\bm{f}_i$ as any of its $\mathcal{A}_i$'s components: $\bm{f}_i\in{\{\bm{f}^a_i, \bm{l}_i, \bm{o}_i\}}\in\mathbb{R}^{D^f}$, where ${D^f}\in\{D^a, 6, 3K\}$ is its respective dimension.
The quantization can be written as,
\begin{equation}
\label{eq:quantize_2}
\begin{aligned}
    \hat{\bm{f}_i} &= \bm{f}_i + \mathcal{U}\left(-\frac{1}{2}, \frac{1}{2}\right)\times \bm{q}_i , \qquad\;\;\; \text{for training} \\
    &= {\rm Round}(\bm{f}_i / \bm{q}_i)\times \bm{q}_i , \qquad \qquad \, \text{for testing}\\
\end{aligned}
\end{equation}
where
\begin{equation} \label{eq:qi}
\begin{aligned}
    \bm{q}_i &= Q_0\times\left(1+{\rm Tanh}\left({\bm{r}}_i\right)\right) \\
    {\bm{r}}_i &= {\rm MLP_q}\left(\bm{f}^h_i\right).
\end{aligned}
\end{equation}
We use a simple MLP-based context model ${\rm MLP_q}$
to predict from hash feature $\bm{f}^h_i$ a refinement ${\bm{r}}_i\in\mathbb{R}$, which is used to adjust the predefined quantization step size $Q_0$. Note that $Q_0$ varies for $\bm{f}^a$, $\bm{l}$, and $\bm{o}$. Eq.~\ref{eq:qi} essentially restricts the quantization step size $\bm{q}_i\in\mathbb{R}$ to be chosen within $\left(0, 2Q_0\right)$, enabling $\hat{\bm{f}}_i$ to closely resemble the original characteristics of $\bm{f}_i$, 
while maintaining a high fidelity.

\begin{figure}[t]
    \centering
    \includegraphics[width=0.93\linewidth]{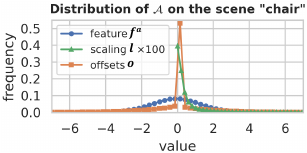}
    \caption{Statistical analysis of the value distributions of $\mathcal{A}$ on the scene ``chair'' of the Synthetic-NeRF dataset~\cite{NeRF}. All three components $\{\bm{f}^a$, $\bm{l}, \bm{o}\}$ exhibit statistical Gaussian distributions. Note that the values of $\bm{l}$ are scaled by a factor of 100 for better visualization.}
    \label{fig:attributes_distribution}
\end{figure}

\noindent\textbf{HAC for Gaussian Distribution Modeling}. To measure the bit consumption of $\hat{\bm{f}}_i$ during training, its probability needs to be estimated in a differentiable manner.
As shown in Fig.~\ref{fig:attributes_distribution}, all three components of anchor attributes $\mathcal{A}$ exhibit statistical tendencies of Gaussian distributions, where $\bm{l}$ displays a single-sided pattern due to Sigmoid activation\footnote{We define $\bm{l}$ as the one \textit{after} Sigmoid activation, which is slightly different from \cite{scaffold}.}, while $\bm{o}$ exhibits an impulse at zero, suggesting the occurrence of substantial redundant Gaussians. 
This observation establishes a lower bound for probability prediction when all $\hat{\bm{f}}_i$s in $\mathcal{A}$ are estimated using the respective $\mu$ and $\sigma$ of the statistical Gaussian Distribution of $\bm{f}^a$, $\bm{l}$ and $\bm{o}$.
Nevertheless, employing a single set of $\mu$ and $\sigma$ for all attributes may lack accuracy. Therefore, we assume anchor attributes $\mathcal{A}$'s values independent, and construct their respective Gaussian distributions, where their individual $\bm{\mu}^\text{s}$ and $\bm{\sigma}^\text{s}$ are estimated by a simple MLP-based context model ${\rm MLP_c}$ from $\bm{f}^h$. 
Specifically, for the $i$th anchor and its quantized anchor attribute vector $\hat{\bm{f}}_i$, with the estimated $\bm{\mu}^\text{s}_i\in\mathbb{R}^D$ and $\bm{\sigma}^\text{s}_i\in\mathbb{R}^D$, we compute the probability of $\hat{\bm{f}}_i$ as, 

\begin{equation}
\begin{aligned}
    p(\hat{\bm{f}}_i) 
    &= \int_{\hat{\bm{f}}_i-\frac{\bm{q}_i}{2}}^{\hat{\bm{f}}_i+\frac{\bm{q}_i}{2}}\phi\left(x \mid \bm{\mu}^\text{s}_i, \bm{\sigma}^\text{s}_i\right)\,dx \\
    &= \Phi(\hat{\bm{f}}_{i} + \frac{\bm{q}_i}{2} \mid \bm{\mu}^\text{s}_i, \bm{\sigma}^\text{s}_i) - \Phi(\hat{\bm{f}}_{i} - \frac{\bm{q}_i}{2} \mid \bm{\mu}^\text{s}_i, \bm{\sigma}^\text{s}_i), \\
    \bm{\mu}^\text{s}_i, \bm{\sigma}^\text{s}_i, \bm{\pi}^\text{s}_i &= {\rm MLP_c}\left(\bm{f}^h_i\right)
\end{aligned}
\label{eq:gaussian_prob}
\end{equation}
where $\phi$ and $\Phi$ represent the probability density function and the cumulative distribution function of Gaussian distribution, respectively.
Note that $\bm{\mu}^\text{s}$ and $\bm{\sigma}^\text{s}$ are Gaussian distribution parameters, and $\bm{\pi}^\text{s}$ represents the weight parameter for distribution combination with the intra-anchor context model that will be introduced in Subsec~\ref{subsec:intra_context}.
This approach effectively captures the relations among anchors, eliminating potential spatial redundancies without compromising the fidelity upper bound of the reconstruction branch.

\subsection{Improving Context Accuracy with Intra Information}
\label{subsec:intra_context}
\noindent\textbf{Intra-Anchor Context Model as Auxiliary Context.} While the proposed HAC context model effectively eliminates spatial redundancies, parameter redundancies still persist within individual anchors. Integrating the intra design from our work FCGS~\cite{FCGS}, we employ an intra-anchor context model to leverage auxiliary information, which further enhances context accuracy.
To construct the intra-anchor context model, we divide each $\bm{f}^a$ into $N^\text{c}$ chunks and employ ${\rm MLP_a}$ to infer the distribution parameters for each chunk based on the previously decoded ones.

\vspace{-10pt}
\begin{equation}
\begin{aligned}
    \bm{\mu}^\text{c}_i, \bm{\sigma}^\text{c}_i, \bm{\pi}^\text{c}_i &= \midoplus_{n^\text{c}=1}^{N^\text{c}}\{\bm{\mu}^\text{c}_{i, n^\text{c}}, \bm{\sigma}^\text{c}_{i, n^\text{c}}, \bm{\pi}^\text{c}_{i, n^\text{c}}\}, \\
    \bm{\mu}^\text{c}_{i, n^\text{c}}, \bm{\sigma}^\text{c}_{i, n^\text{c}}, \bm{\pi}^\text{c}_{i, n^\text{c}} &= {\rm MLP_a}([\hat{\bm{f}}^a_{i, [0,n^\text{c}c-c)}; \bm{\mu}^\text{s}_i; \bm{\sigma}^\text{s}_i; \bm{\pi}^\text{s}_i])
\end{aligned}
\end{equation}
where $n^\text{c}$ and $c$ denote the chunk index and the number of channels per chunk, respectively. $\bm{\mu}^\text{c}_{i, n^\text{c}}, \bm{\sigma}^\text{c}_{i, n^\text{c}}, \bm{\pi}^\text{c}_{i, n^\text{c}}$ represent the probability parameters for chunk $n^\text{c}$, each with a dimensional size of $c$. $\midoplus$ is the channel-wise concatenate operation. $\bm{\pi}^\text{c}$ is the weight parameter for distribution combination with HAC in GMM. It is important to note that this approach is applied only to the anchor feature $\bm{f}^a$, as the internal redundancies of other attributes are negligible. Specifically, for the scaling $\bm{l}$, it only has a dimensionality of $6$, making the potential storage savings from reducing its internal redundancies minimal. Moreover, achieving such reduction requires additional MLPs, introducing extra storage and coding complexity. As for the offset $\bm{o}$, the adaptive offset masking strategy introduced in Subsec.~\ref{subsec:adaptive_offset_mask} has already removed redundant Gaussians. Further attempts to reduce redundancies among the ``$xyz$'' coordinates using the intra-anchor context are challenging. Detailed results can be found in the ablation studies in Subsec.~\ref{subsec:ablation}.

\noindent\textbf{Gaussian Mixture Model (GMM)}.
To achieve joint probability estimation for the anchor feature $\bm{f}^a$ using both context models (\ie, HAC and the intra-anchor context model), we employ GMM. This approach combines the Gaussian distribution parameter sets from each context model to represent the final probability. The probability estimation is formulated as follows:

\vspace{-10pt}
\begin{small}
\begin{equation}
\begin{aligned}
    p(\hat{\bm{f}}^a_{i}) &= \sum_{l\in{\{\text{s}, \text{c}\}}}{\bm{\theta}}_{i}^l\left(\Phi(\hat{\bm{f}}^a_{i} + \frac{\bm{q}_i}{2} \mid {\bm{\mu}}_{i}^l, {\bm{\sigma}}_{i}^l) - \Phi(\hat{\bm{f}}^a_{i} - \frac{\bm{q}_i}{2} \mid {\bm{\mu}}_{i}^l, {\bm{\sigma}}_{i}^l)\right), \\
    {\bm{\theta}}_{i}^l &= \frac{\exp({\bm{\pi}}_{i}^l)}{\sum_{v\in{\{\text{s}, \text{c}\}}} \exp({\bm{\pi}}_{i}^v)}
\end{aligned}
\end{equation}
\end{small}

This formulation ensures that the GMM adaptively weights the contributions of the two context models for probability estimation. By dynamically balancing their respective distribution parameter sets, it yields a more accurate and robust probability estimation for the anchor feature $\bm{f}^a$.

\subsection{Adaptive Offset Masking}
\label{subsec:adaptive_offset_mask}
From Fig.~\ref{fig:attributes_distribution}, we can observe that the offset $\bm{o}$ exhibits an impulse at zero, suggesting the occurrence of substantial unnecessary Gaussians and subsequently, anchors. 

Pruning trivial Gaussians or anchors directly reduces the number of parameters, thereby slimming the model. To this end, we draw inspiration from~\cite{Lee}, which employs learnable binary masks updated via the straight-through~\cite{STE} strategy to eliminate invalid Gaussians. Notably, it utilizes an additional loss term (\ie, $L_m$) to regularize the mask rate, which balances the degree of compression.
However, directly applying this approach to our HAC++ framework by incorporating an extra mask loss term poses challenges. Specifically, the weight of $L_m$ must be manually adjusted across different RD trade-off points in Eq.~\ref{eq:loss} as $\lambda$ varies to achieve optimal mask ratios. This process is tedious, and difficult to optimize effectively.

To address this problem, we incorporate the mask into the rate calculation, allowing the mask ratio to be adaptively adjusted according to the rate via backpropagation. Firstly, following~\cite{Lee}, for each anchor, the Gaussian-level mask $\bm{m} \in \mathbb{R}^K$ is obtained as:

\begin{equation}
    \bm{m}_i = \text{sg}\left(\mathds{1}[\texttt{Sig}(\bm{f}^m_i) > \epsilon_m] - \texttt{Sig}(\bm{f}^m_i)\right) + \texttt{Sig}(\bm{f}^m_i) \\
\end{equation}
where $\bm{f}^m \in \mathbb{R}^K$ is a learnable feature to deduce the mask, $\texttt{Sig}$ represents the sigmoid function, and $\text{sg}$ is the stop-gradient operator. A value of $1$ in $\bm{m}$ indicates the corresponding Gaussian offset is valid, while $0$ denotes invalid. Invalid Gaussian offsets can be removed to save parameters. Furthermore, if all offsets associated with an anchor are pruned, then the anchor becomes irrelevant for rendering and can be entirely removed, including its $\bm{x}^a$, $\mathcal{A}$, and $\bm{m}$. To make the model aware of the rate change by anchor pruning, we further introduce an anchor-level mask $\bm{m}^a \in \mathbb{R}$, which is derived from the Gaussian-level mask $\bm{m}$. This design encourages anchor pruning, thereby enhancing parameter savings:

\begin{equation}
    m^a_i = \text{sg}\left(\mathds{1}[\overline{m}_i > 0] - \overline{m}_i\right) + \overline{m}_i, \; \overline{m}_i = \frac{1}{K}\sum_{k=1}^K m_{i, k}
    \label{eq:mask_a}
\end{equation}
where $\overline{m}$ represents the average offset mask ratio of an anchor. If all the offsets are pruned on an anchor (\ie, $\overline{m}=0$), then this anchor no longer contributes to rendering and should be pruned entirely (including its $\bm{x}^a$, $\mathcal{A}$ and $\bm{m}$).

To enable adaptive mask updates, the mask information should be involved into both the rendering process and entropy estimation in a differentiable manner. For the rendering process, the Gaussian-level mask is applied to the opacity and scale of each Gaussian as $m_{i,k}\alpha_{i,k}$ and $m_{i,k}s_{i,k}$, ensuring invalid Gaussians do not contribute to rendering while valid ones remain unaffected. For entropy estimation, both Gaussian-level and anchor-level masks are included in the bit consumption calculation, making the model explicitly aware of the pruning scheme of both Gaussians and anchors. The overall entropy loss is then defined as the bit consumption $b$ across all anchors:

\begin{equation}
\begin{aligned}
    L_{\text{entropy}} = \;
    & \sum_i^N b_i, \; \text{where} \\
    b_i = \;
    & m^a_i\sum_{\bm{f}\in\{\bm{f}^a, \bm{l}\}} \sum_{j=1}^{D^f} \left(-\log_2 p(\hat{{f}}_{i,j})\right) + \\
    & m^a_i\sum_{\bm{f}\in\{\bm{o}\}} \sum_{k=1}^{K}m_{i,k}\sum_{j=1}^{3} \left(-\log_2 p(\hat{{f}}_{i,3k+j})
    \right)
\end{aligned}
\end{equation}
where $N$ is the total number of anchors, and $b_i$ indicates the bit consumption of the $i$-th anchor. Minimizing $L_{\text{entropy}}$ promotes accurate probability estimation by $p(\hat{\bm{f}}_i)$, which in turn guides the context models' learning. By incorporating mask information into both paths' gradient chain, this approach ensures adaptive updates across different $\lambda$ constraints in Eq.~\ref{eq:loss}, eliminating the need for additional loss terms. Consequently, the mask update process dynamically identifies the optimal pruning ratio.

\subsection{Hash Grid Compression}
As shown in~\ref{eq:prior_probability}, the size of the hash grid $\mathcal{H}$ also significantly influences the final storage size. To this end, we binarize the hash table to $\{-1, +1\}$ using straight-through estimation (STE)~\cite{BiRF} and calculate the occurrence frequency $h_f$~\cite{cnc2024} of the symbol ``$+1$'' to estimate its bit consumption:
\begin{equation}
    L_{\text{hash}} = M_+\times\left(-\log_2(h_f)\right) + M_-\times\left(-\log_2(1-h_f)\right)
\end{equation}
where $M_+$ and $M_-$ are total numbers of ``$+1$'' and ``$-1$'' in the hash grid.

\subsection{Training and Coding Process}
During training, we incorporate both the rendering fidelity loss and the entropy loss to ensure the model improves rendering quality while controlling total bitrate consumption in a differentiable manner. Our overall loss is
\begin{equation}
    Loss = L_{\text{Scaffold}} + \lambda\frac{1}{N(D^a+6+3K)}(L_{\text{entropy}} + L_{\text{hash}}).
    \label{eq:loss}
\end{equation}
Here, $L_{\text{Scaffold}}$ represents the rendering loss as defined in~\cite{scaffold}, which includes two fidelity penalty loss terms and one regularization term for the scaling $\bm{l}$. The second part in Eq.~\ref{eq:loss} is the estimated controllable bit consumption, including the estimated bits $L_{\text{entropy}}$ for anchor attributes and $L_{\text{hash}}$ for the hash grid.
$\lambda$ is the trade-off hyperparameters used to balance the rate and fidelity.

For the encoding/decoding process, the anchor location $\bm{x}^a$ and the binary hash grid $\mathcal{H}$ are initially encoded/decoded separately using Geometric Point Cloud Compression (GPCC)~\cite{GPCC} and AE with $h_f$, respectively. Then, hash feature $\bm{f}^h$ is obtained through interpolation based on $\mathcal{H}$ and $\bm{x}^a$. Once $\bm{f}^h$ is acquired, the context models ${\rm MLP_q}$ and ${\rm MLP_c}$ are then employed to estimate quantization refinement term $\bm{r}$ and parameters of the Gaussian Distribution (\ie, $\bm{\mu}^\text{s}$ and $\bm{\sigma}^\text{s}$) to derive the probability $p(\bm{\hat{f}})$ for entropy encoding/decoding with AE. For the anchor feature $\bm{f}^a$, the intra-anchor context model is also integrated to enhance the overall context accuracy via the GMM. Consequently, it is encoded/decoded sequentially chunk by chunk.

\section{Experiments}
In this section, we first outline the implementation details of our HAC++ framework (Subsec.~\ref{subsec:implementation}) and evaluate its performance compared to existing 3DGS compression methods (Subsec.~\ref{subsec:experiment}). Additionally, we conduct ablation studies to demonstrate the effectiveness of each technical component (Subsec.~\ref{subsec:ablation}). Moreover, Subsec.~\ref{subsec:mask_ratio} presents the variation in mask ratios across different RD trade-off points. To further illustrate the role of context models, we visualize the bit allocation map (Subsec.~\ref{subsec:visual_bit}). Finally, we present in-depth statistical analyses of HAC++ from various perspectives, including storage size (Subsec.~\ref{subsec:decom_size}), coding time (Subsec.~\ref{subsec:decom_coding}), training and inference efficiency (Subsec.~\ref{subsec:train_render_efficiency}), and performance variations across training iterations (Subsec.~\ref{subsec:training_diff_iter}).

\subsection{Implementation Details}
\label{subsec:implementation}
\textbf{Basic Settings.} We implement our HAC++ method based on the Scaffold-GS repository~\cite{scaffold} using the PyTorch framework~\cite{pytorch} and train the model on a single NVIDIA L40s GPU with $48$ GB memory. 
We increase the dimension of the Scaffold-GS anchor feature $\bm{f}^a$ (\ie, $D^a$) to 50, and disable its feature bank as we found it may lead to unstable training. Other hyperparameters remain unchanged (\eg, the number of offsets per anchor $K=10$). For the hash grid $\mathcal{H}$, we utilize a mixed 3D-2D structured binary hash grid, with $12$ levels of 3D embeddings ranging from $16$ to $512$ resolutions, and $4$ levels of 2D embeddings ranging from $128$ to $1024$ resolutions. The maximum hash table sizes are $2^{13}$ and $2^{15}$ for the 3D and 2D grids, respectively, both with a feature dimension of $D^h=4$. We change $\lambda$ from $0.5e-3$ to $4e-3$ to achieve variable bitrates. We set $Q_0$ as $1$, $0.001$ and $0.2$ for $\bm{f}^a$, $\bm{l}$ and $\bm{o}$, respectively. The number of intra steps is $N^\text{c}=5$.
In implementation, ${\rm MLP_q}$ and ${\rm MLP_c}$ are combined into a single 3-layer MLP with ReLU activation, while ${\rm MLP_a}$ consists of multiple MLPs with varying input channels to accommodate different number of input chunks across different intra steps. For the Synthetic-NeRF~\cite{NeRF} dataset, a lightweight version of ${\rm MLP_a}$ is employed to minimize overhead. Note that the total training iteration of HAC++ is $30k$, which is consistent with Scaffold-GS and 3DGS methods to ensure a fair comparison.

\noindent\textbf{Sampling Strategy}.
During training, using all anchors for entropy training in each iteration results in prolonged training time and potential out-of-memory (OOM) issues. Therefore, we adopt a sampling strategy: in each iteration, we only randomly sample and entropy train 5\% anchors from all.

\begin{figure}[t]
    \centering
    \includegraphics[width=1.00\linewidth]{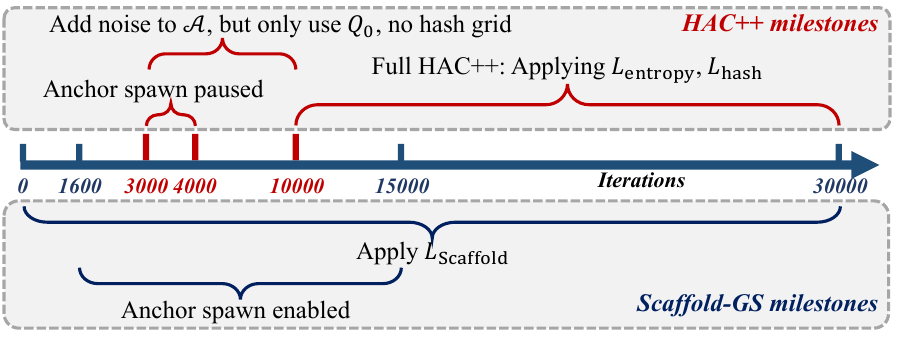}
    \caption{Detailed training process of HAC++. We use \textbf{\textcolor{supple_red}{red lines}} and \textbf{\textcolor{supple_blue}{blue lines}} to indicate the training process of our model and Scaffold-GS, respectively.}
    \label{fig:training_process}
\end{figure}

\noindent\textbf{Training Process.}
We enhance the training process to improve stability, as shown in Fig.~\ref{fig:training_process}. \textit{During the initial $3k$ iterations}, we train the original Scaffold-GS~\cite{scaffold} model to ensure a stable start of the anchor attribute training and anchor spawning process. \textit{From iteration $3k$ to $10k$}, we add noise to anchor attributes $\mathcal{A}$, which allows the model to adapt to quantization. Note that, in this stage, we only apply $Q_0$ for quantization without using $\bm{r}$. Therefore, we do not need the hash grid. Specifically, we pause the anchor spawning process between iterations $3k$ and $4k$ for a transitional period, as the sudden introduction of quantization may introduce instability to the spawning process. \textit{After iteration $10k$}, assuming the 3D model is fitted to quantization, we fully integrate the HAC++ framework to jointly train the context models.

\begin{table*}[t]
    \centering
    \caption{Quantitative Results. 3DGS~\cite{3DGS} and Scaffold-GS~\cite{scaffold} Are Two Baselines. For Our Approach, We Provide Two Results with Different Size and Fidelity Trade-Offs by Adjusting $\lambda$. A Smaller $\lambda$ Results in a Larger Size but Improved Fidelity, and Vice Versa. The Best and Second-Best Results Are Highlighted in \colorbox{red!25}{Red} and \colorbox{yellow!25}{Yellow} Cells. The Sizes Are Measured in MB.
    }
    \begin{tabular}{ll|cccc|cccc|cccc}
        \toprule
        \multicolumn{2}{l|}{\textbf{Datasets}}          & \multicolumn{4}{c|}{\textbf{Synthetic-NeRF~\cite{NeRF}}} & \multicolumn{4}{c|}{\textbf{Mip-NeRF360~\cite{mip360}}} & \multicolumn{4}{c}{\textbf{Tank\&Temples~\cite{tant}}} \\
        \multicolumn{2}{l|}{\textbf{methods}} & psnr$\uparrow$    & ssim$\uparrow$   & lpips$\downarrow$ & size$\downarrow$   & psnr$\uparrow$    & ssim$\uparrow$   & lpips$\downarrow$ & size$\downarrow$   & psnr$\uparrow$   & ssim$\uparrow$   & lpips$\downarrow$ & size$\downarrow$   \\
        \bottomrule
        \multicolumn{2}{l|}{\textbf{3DGS~\cite{3DGS}}}&\cellcolor{red!25}{33.80}&\cellcolor{red!25}{0.970}&\cellcolor{red!25}{0.031}&68.46&27.46&\cellcolor{red!25}{0.812}&\cellcolor{red!25}{0.222}&750.9&23.69&0.844&0.178&431.0    \\
        \multicolumn{2}{l|}{\textbf{Scaffold-GS~\cite{scaffold}}}&33.41&0.966&0.035&19.36&27.50&0.806&0.252&253.9&23.96&0.853&\cellcolor{yellow!25}{0.177}&86.50    \\  \hline
        \multicolumn{2}{l|}{\textbf{Lee~\etal~\cite{Lee}}}&33.33&0.968&0.034&5.54&27.08&0.798&0.247&48.80&23.32&0.831&0.201&39.43    \\  
        \multicolumn{2}{l|}{\textbf{Compressed3D~\cite{Simon}}}&32.94&0.967&\cellcolor{yellow!25}{0.033}&3.68&26.98&0.801&0.238&28.80&23.32&0.832&0.194&17.28    \\  
        \multicolumn{2}{l|}{\textbf{EAGLES\cite{EAGLES}}}&32.51&0.964&0.039&4.26&27.14&0.809&0.231&58.91&23.28&0.835&0.203&28.99    \\
        \multicolumn{2}{l|}{\textbf{LightGaussian~\cite{Lightgaussian}}}&32.73&0.965&0.037&7.84&27.00&0.799&0.249&44.54&22.83&0.822&0.242&22.43    \\  
        \multicolumn{2}{l|}{\textbf{SOG~\cite{SOG}}}&31.37&0.959&0.043&2.00&26.56&0.791&0.241&16.70&23.15&0.828&0.198&9.30    \\
        \multicolumn{2}{l|}{\textbf{Navaneet~\etal~\cite{Navaneet}}}&32.99&0.966&0.037&3.10&27.12&0.806&0.240&19.33&23.44&0.838&0.198&12.50    \\
        \multicolumn{2}{l|}{\textbf{Reduced3DGS~\cite{Reduced3DGS}}}&33.02&0.967&0.035&2.11&27.19&0.807&\cellcolor{yellow!25}{0.230}&29.54&23.57&0.840&0.188&14.00    \\
        \multicolumn{2}{l|}{\textbf{RDOGaussian~\cite{RDOGaussian}}}&33.12&0.967&0.035&2.31&27.05&0.802&0.239&23.46&23.34&0.835&0.195&12.03    \\
        \multicolumn{2}{l|}{\textbf{MesonGS-FT~\cite{MesonGS}}}&32.92&0.968&\cellcolor{yellow!25}{0.033}&3.66&26.99&0.796&0.247&27.16&23.32&0.837&0.193&16.99    \\  \hline
        \multicolumn{2}{l|}{\textbf{HAC (lowrate)~\cite{HAC}}}&33.24&0.967&0.037&1.18&27.53&0.807&0.238&15.26&24.04&0.846&0.187&8.10    \\  
        \multicolumn{2}{l|}{\textbf{HAC (highrate)~\cite{HAC}}}&33.71&0.968&0.034&1.86&\cellcolor{yellow!25}{27.77}&\cellcolor{yellow!25}{0.811}&\cellcolor{yellow!25}{0.230}&21.87&\cellcolor{red!25}{24.40}&0.853&\cellcolor{yellow!25}{0.177}&11.24    \\
        \multicolumn{2}{l|}{\textbf{ContextGS (lowrate)~\cite{Contextgs}}}&32.79&0.965&0.040&\cellcolor{yellow!25}{1.01}&27.62&0.808&0.237&12.68&24.20&0.852&0.184&7.05    \\  
        \multicolumn{2}{l|}{\textbf{ContextGS (highrate)~\cite{Contextgs}}}&33.51&0.968&0.035&1.56&27.75&\cellcolor{yellow!25}{0.811}&0.231&18.41&24.29&\cellcolor{red!25}{0.855}&\cellcolor{red!25}{0.176}&11.80    \\
        \multicolumn{2}{l|}{\textbf{CompGS (lowrate)~\cite{liu2024compgs}}}&/&/&/&/&26.37&0.778&0.276&\cellcolor{yellow!25}{8.83}&23.11&0.815&0.236&\cellcolor{yellow!25}{5.89}    \\
        \multicolumn{2}{l|}{\textbf{CompGS (highrate)~\cite{liu2024compgs}}}&/&/&/&/&27.26&0.803&0.239&16.50&23.70&0.837&0.208&9.60    \\  \hline
        \multicolumn{2}{l|}{\textbf{Ours HAC++ (lowrate)}}&33.03&0.966&0.039&\cellcolor{red!25}{0.88}&27.60&0.803&0.253&\cellcolor{red!25}{8.34}&24.22&0.849&0.190&\cellcolor{red!25}{5.18}    \\  
        \multicolumn{2}{l|}{\textbf{Ours HAC++ (highrate)}}&\cellcolor{yellow!25}{33.76}&\cellcolor{yellow!25}{0.969}&\cellcolor{yellow!25}{0.033}&1.84&\cellcolor{red!25}{27.82}&\cellcolor{yellow!25}{0.811}&0.231&18.48&\cellcolor{yellow!25}{24.32}&\cellcolor{yellow!25}{0.854}&0.178&8.63    \\
        \toprule
    \end{tabular}
    \centering
    \begin{tabular}{ll|cccc|cccc}
        \toprule
        \multicolumn{2}{l|}{\textbf{Datasets}} & \multicolumn{4}{c|}{\textbf{DeepBlending~\cite{deepblending}}} & \multicolumn{4}{c}{\textbf{BungeeNeRF~\cite{BungeeNeRF}}} \\
        \multicolumn{2}{l|}{\textbf{methods}} & psnr$\uparrow$    & ssim$\uparrow$   & lpips$\downarrow$ & size$\downarrow$   & psnr$\uparrow$    & ssim$\uparrow$   & lpips$\downarrow$ & size$\downarrow$   \\
        \bottomrule
        \multicolumn{2}{l|}{\textbf{3DGS~\cite{3DGS}}}&29.42&0.899&\cellcolor{red!25}{0.247}&663.9&24.87&0.841&0.205&1616    \\
        \multicolumn{2}{l|}{\textbf{Scaffold-GS~\cite{scaffold}}}&30.21&0.906&0.254&66.00&26.62&0.865&0.241&183.0    \\  \hline
        \multicolumn{2}{l|}{\textbf{Lee~\etal~\cite{Lee}}}&29.79&0.901&0.258&43.21&23.36&0.788&0.251&82.60    \\  
        \multicolumn{2}{l|}{\textbf{Compressed3D~\cite{Simon}}}&29.38&0.898&0.253&25.30&24.13&0.802&0.245&55.79    \\  
        \multicolumn{2}{l|}{\textbf{EAGLES\cite{EAGLES}}}&29.72&0.906&\cellcolor{yellow!25}{0.249}&52.34&25.89&0.865&\cellcolor{yellow!25}{0.197}&115.2    \\
        \multicolumn{2}{l|}{\textbf{LightGaussian~\cite{Lightgaussian}}}&27.01&0.872&0.308&33.94&24.52&0.825&0.255&87.28    \\  
        \multicolumn{2}{l|}{\textbf{SOG~\cite{SOG}}}&29.12&0.892&0.270&5.70&22.43&0.708&0.339&48.25   \\
        \multicolumn{2}{l|}{\textbf{Navaneet~\etal~\cite{Navaneet}}}&29.90&0.907&0.251&13.50&24.70&0.815&0.266&33.39    \\
        \multicolumn{2}{l|}{\textbf{Reduced3DGS~\cite{Reduced3DGS}}}&29.63&0.902&\cellcolor{yellow!25}{0.249}&18.00&24.57&0.812&0.228&65.39    \\
        \multicolumn{2}{l|}{\textbf{RDOGaussian~\cite{RDOGaussian}}}&29.63&0.902&0.252&18.00&23.37&0.762&0.286&39.06    \\
        \multicolumn{2}{l|}{\textbf{MesonGS-FT~\cite{MesonGS}}}&29.51&0.901&0.251&24.76&23.06&0.771&0.235&63.11    \\  \hline
        \multicolumn{2}{l|}{\textbf{HAC (lowrate)~\cite{HAC}}}&29.98&0.902&0.269&4.35&26.48&0.845&0.250&18.49    \\  
        \multicolumn{2}{l|}{\textbf{HAC (highrate)~\cite{HAC}}}&\cellcolor{yellow!25}{30.34}&0.906&0.258&6.35&27.08&0.872&0.209&29.72    \\
        \multicolumn{2}{l|}{\textbf{ContextGS (lowrate)~\cite{Contextgs}}}&30.11&0.907&0.265&\cellcolor{yellow!25}{3.45}&26.90&0.866&0.222&\cellcolor{yellow!25}{14.00}    \\  
        \multicolumn{2}{l|}{\textbf{ContextGS (highrate)~\cite{Contextgs}}}&\cellcolor{red!25}{30.39}&\cellcolor{yellow!25}{0.909}&0.258&6.60&\cellcolor{yellow!25}{27.15}&\cellcolor{yellow!25}{0.875}&0.205&21.80    \\
        \multicolumn{2}{l|}{\textbf{CompGS (lowrate)~\cite{liu2024compgs}}}&29.30&0.895&0.293&6.03&/&/&/&/    \\
        \multicolumn{2}{l|}{\textbf{CompGS (highrate)~\cite{liu2024compgs}}}&29.69&0.901&0.279&8.77&/&/&/&/    \\ \hline
        \multicolumn{2}{l|}{\textbf{Ours HAC++ (lowrate)}}&30.16&0.907&0.266&\cellcolor{red!25}{2.91}&26.78&0.858&0.235&\cellcolor{red!25}{11.75}    \\  
        \multicolumn{2}{l|}{\textbf{Ours HAC++ (highrate)}}&\cellcolor{yellow!25}{30.34}&\cellcolor{red!25}{0.911}&0.254&5.28&\cellcolor{red!25}{27.17}&\cellcolor{red!25}{0.879}&\cellcolor{red!25}{0.196}&20.82    \\
        \toprule
    \end{tabular}
    \label{tab:main_quantitative}
    \vspace{-15pt}
\end{table*}

\subsection{Experiment Evaluation}
\label{subsec:experiment}
\textbf{Baselines}.
We compare HAC++ against a wide range of existing 3DGS compression methods. Approaches such as~\cite{Lee, Simon, Navaneet, Lightgaussian, Reduced3DGS, RDOGaussian, MesonGS} primarily rely on codebook-based or parameter pruning techniques. EAGLES~\cite{EAGLES} and SOG~\cite{SOG} apply entropy constraints and sorting strategies to minimize storage, respectively. For MesonGS~\cite{MesonGS}, we utilize its finetuned variant to achieve improved RD performance. Additionally, Scaffold-GS~\cite{scaffold} introduces anchor points for a compact representation. Building on Scaffold-GS, HAC~\cite{HAC}, CompGS~\cite{liu2024compgs} and ContextGS~\cite{Contextgs} incorporate context models to further reduce model size.

\noindent\textbf{Datasets}. We follow Scaffold-GS to perform evaluations on multiple datasets, including a small-scale Synthetic-NeRF~\cite{NeRF} and four large-scale real-scene datasets: BungeeNeRF~\cite{BungeeNeRF}, DeepBlending~\cite{deepblending}, Mip-NeRF360~\cite{mip360}, and Tanks\&Temples~\cite{tant}. Note that we evaluate the entire 9 scenes from Mip-NeRF360 dataset~\cite{mip360}. Covering diverse scenarios, these datasets allow us to comprehensively demonstrate the effectiveness of all methods.

\noindent\textbf{Metrics}. To comprehensively evaluate compression RD performance, we calculate relative rate (size) change of our approach over others under a similar fidelity. We further provide BD-rate~\cite{BD_rate} in our ablation studies to better reflect performance changes.

\begin{figure*}[t]
    \centering
    \includegraphics[width=0.95\linewidth]{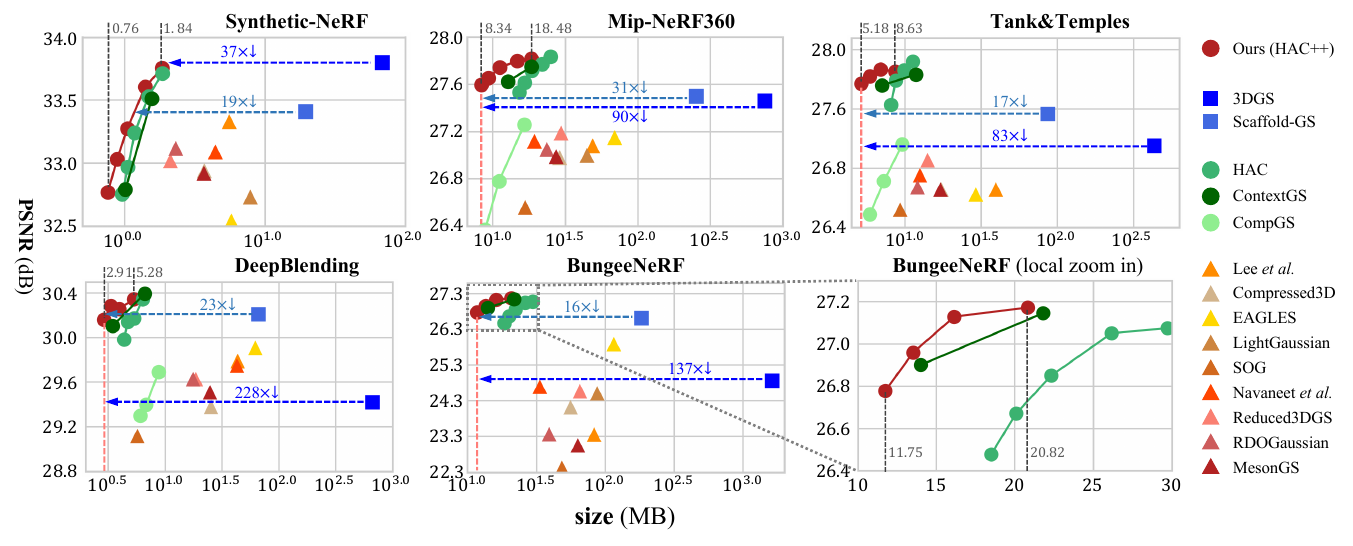}
    \caption{RD curves for quantitative comparisons. We vary $\lambda$ to achieve variable bitrates. Note that ${\rm log_{10}}$ scale is used for x-axis for better visualization.
    }
    \label{fig:main_curve}
\end{figure*}

\begin{figure*}[t]
    \centering
    \includegraphics[width=0.85\linewidth]{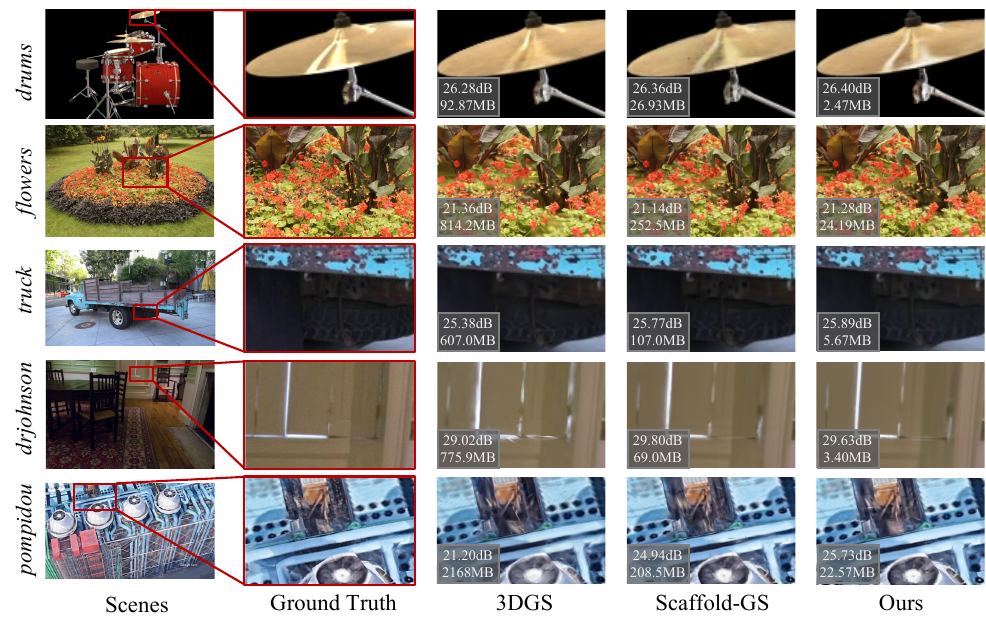}
    \caption{Qualitative comparisons of scenes across different datasets. PSNR and size results are given at lower-left.}
    \label{fig:main_qualitative}
\end{figure*}

\noindent\textbf{Results}.
Quantitative results are shown in~\ref{tab:main_quantitative} and~\ref{fig:main_curve}, the qualitative outputs are illustrated in~\ref{fig:main_qualitative}. Full per-scene results of HAC++ across multiple fidelity metrics (\ie, PSNR, SSIM~\cite{ssim}, LPIPS~\cite{LPIPS}, and size) can be found in Sec.~\ref{sec:per_scene}. HAC++ achieves significant size reductions, surpassing $100\times$ compared to the vanilla 3DGS~\cite{3DGS} on average across all datasets, while also delivering improved fidelity. Moreover, it achieves over $20\times$ size reduction compared to the base model, Scaffold-GS~\cite{scaffold}. Notably, HAC++ surpasses Scaffold-GS in fidelity, primarily due to two factors:
1) the entropy loss effectively regularizes the model to prevent overfitting, and 
2) we increase the dimension of the anchor feature (\ie, $D^a$) to $50$, resulting in a larger model volume.
For methods primarily relying on codebooks and pruning techniques (mid-chunk), their designs struggle to achieve significant storage reductions due to the limited utilization of contextual information among parameters. While SOG~\cite{SOG} achieves a small size, it significantly sacrifices fidelity. While ContextGS~\cite{Contextgs} and CompGS~\cite{liu2024compgs} each introduce their context models, HAC++ demonstrates superior performance owing to its more effective and well-optimized context model designs and mask strategies.

\noindent\textbf{Bitstream}.
The bitstream of HAC++ comprises five components: anchor attributes $\mathcal{A}$ (\ie, $\bm{f}^a$, $\bm{l}$, and $\bm{o}$), binary hash grid $\mathcal{H}$, offset masks $\bm{m}$, anchor locations $\bm{x}^a$, and MLP parameters. Among these, $\mathcal{A}$ is entropy-encoded using AE~\cite{AE}, leveraging probabilities estimated by the context models, and constitutes the dominant portion of storage. The hash grid $\mathcal{H}$ and masks $\bm{m}$ are binary data encoded by AE based on their occurrence frequencies. Anchor locations $\bm{x}^a$ are 16-bit quantized and losslessly compressed using GPCC~\cite{GPCC}. The MLP parameters are stored directly using $32$-bit precision. For detailed analysis of storage size and coding time, please refer to Subsec.~\ref{subsec:decom_size} and~\ref{subsec:decom_coding}, respectively.

\begin{table*}[t]
    \centering
    \normalsize
    \caption{Ablation Studies on the Mip-NeRF360 Dataset~\cite{mip360}. Positive BD-Rate Values Indicate Increased Sizes Compared to Anchor Method HAC++ at the Same Fidelity, Which are Unexpected.}
    \begin{tabular}{l|c|cc}
        \toprule[2pt]
        \textbf{Ablation Item}                  &     \textbf{BD-rate} $\downarrow$ &     \textbf{Train Time} (s) &   \textbf{FPS} \\ \toprule
        W/o AQM                                 &    N/A   &    2073    &    140    \\  \hline
        W/o HAC information                     &    $+63.3\%$  &    2303   &    144      \\
        W/o intra-anchor probability            &    $+14.7\%$   &    2130   &    143    \\ 
        W/o using GMM for probability fusion         &    $+5.7\%$  &    2289   &    140      \\
        W/ intra-anchor context on the scaling  $\bm{l}$        &    $-0.1\%$   &    2386  &    142    \\
        W/ intra-anchor context on the offset  $\bm{o}$        &    $+0.9\%$   &    2456   &    141    \\ \hline
        W/o adaptive offset masking $\bm{m}$          &    $+31.4\%$    &    2239   &    125   \\
        W/o anchor-level mask $\bm{m}^a$ in entropy loss  &   $+9.3\%$   &   2277    &    137    \\
        W/ extra mask loss term, instead of mask-aware rate  &    $+9.6\%$   &    2208    &    147      \\ \hline
        W/o using GPCC for location coding                 &    $+14.0\%$   &    2292   &    141     \\ \hline
        \textbf{HAC++} (anchor method)                             &     $0.0\%$   &     2292  &   141    \\   \bottomrule[2pt]
    \end{tabular}
    \label{tab:ablation}
    \vspace{-15pt}
\end{table*}

\begin{figure}[t]
    \hspace*{-1cm}
    \centering
    \includegraphics[width=0.90\linewidth]{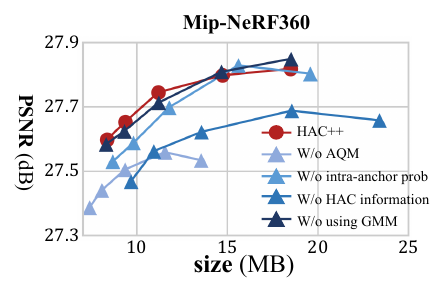}
    \caption{RD curve of ablation study on \textbf{context models}. Experiments are conducted on the Mip-NeRF360 dataset~\cite{mip360} dataset. We vary $\lambda$ for variable rates.}
    \label{fig:ablation}
\end{figure}

\begin{figure}[t]
    \centering
    \includegraphics[width=1.0\linewidth]{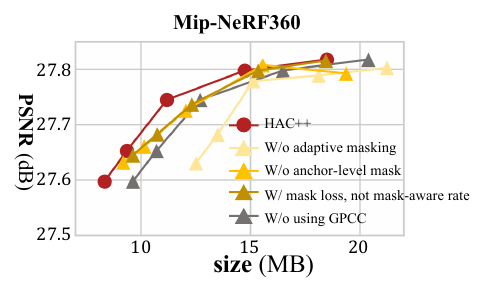}
    \caption{RD curve of ablation study on \textbf{masking strategies} and \textbf{GPCC}. Experiments are conducted on the Mip-NeRF360 dataset~\cite{mip360}. We vary $\lambda$ for variable rates.}
    \label{fig:ablation_mask}
\end{figure}

\subsection{Ablation Study}
\label{subsec:ablation}
In this subsection, we conduct ablation studies to evaluate the effectiveness of individual technical components in HAC++. Our experiments are performed on the Mip-NeRF360 dataset~\cite{mip360}, which features large-scale and diverse scenes, offering reliable results. Quantitative comparisons are presented as curves in Fig.~\ref{fig:ablation} and~\ref{fig:ablation_mask}. BD-rate~\cite{BD_rate} is calculated to measure the relative size change compared to the full HAC++ method at the same fidelity in Tab.~\ref{tab:ablation}. Note that positive BD-rate values indicate increased sizes at the same fidelity compared to the anchor method (\ie, HAC++), which is undesirable. The ablation studies encompass three aspects:

\begin{itemize}
    \item \textbf{Fidelity Preservation}. (Fig.~\ref{fig:ablation} and Tab.~\ref{tab:ablation})
    We investigate the impact of disabling the adaptive quantization module (AQM). Specifically, the adaptive term $\bm{r}$ is removed, retaining only $Q_0$ to ensure a necessary decimal quantization step. This modification leads to a significant drop in fidelity, especially at higher rates, as the anchor attributes $\mathcal{A}$ fail to retain sufficient information for rendering after quantization. The drastic reduction in fidelity results in an incalculable BD-rate due to the absence of overlap with HAC++ in fidelity, which is denoted as N/A in Tab.~\ref{tab:ablation}.
    \item \textbf{Probability Estimation Accuracy}. (Fig.~\ref{fig:ablation} and Tab.~\ref{tab:ablation}) To assess context models, we perform three experiments: \textbf{(1)} We set the hash grid to all zeros to eliminate mutual information from HAC. This downgrades the conditional probability from $p(\mathcal{A}|\bm{f}^h)$ to $p(\mathcal{A})$, leading to inaccurate probability estimation, consequently, a significantly larger model size. \textbf{(2)} We remove the auxiliary intra-anchor context model, which also causes a noticeable increase in mode size. Here’s a refined version for clarity and style:
    \textbf{(3)} To evaluate the approach of fusing the two context models, we examine the effectiveness of the GMM. Specifically, we replace the GMM with a simpler concatenation strategy for combining the two context models. This modification results in only a \textit{single} set of Gaussian distribution parameters being deduced for probability estimation.
    As observed from the 5th line in Tab.~\ref{tab:ablation}, this approach is suboptimal, since the GMM provides a more flexible distribution estimation to better approximate the true conditional distribution. We also evaluate the impact of applying the intra-anchor context model to the scaling $\bm{l}$ and the offset $\bm{o}$ individually. The results in Tab.~\ref{tab:ablation} show negligible improvements. As their RD curves overlap significantly with that of HAC++, we omit them from Fig.~\ref{fig:ablation} for clarity. As highlighted in Tab.~\ref{tab:ablation}, enhancing entropy estimation accuracy contributes to an improved BD-rate while preserving comparable training and rendering efficiency.
    \item \textbf{Masking Strategies}. (Fig.~\ref{fig:ablation} and Tab.~\ref{tab:ablation}) We evaluate the effects of different mask strategies, including Gaussian-level and anchor-level masks. The Gaussian-level mask $\bm{m}$, which is applied consistently to both rendering and entropy paths, demonstrates its effectiveness in reducing parameter count and enhancing compression performance. For the anchor-level mask $\bm{m}^a$, its removal makes the anchor masking scheme passive, reducing its effectiveness in eliminating redundant anchors. We also present the result of using an extra mask loss term for regularization as an alternative to the proposed mask-aware rate calculation, which proves less effective. Finally, employing GPCC for anchor location coding successfully reduces the storage size further. As shown in Tab.~\ref{tab:ablation}, pruning invalid Gaussians and anchors not only enhances compression performance but also improves rendering efficiency. 
\end{itemize}

Overall, each technical component in HAC++ contributes to improved rate-distortion performance. Collectively, these components form a robust framework for effective 3DGS compression.

\begin{table*}[t]
    \small
    \centering
    \caption{\textbf{Encoding Time} for Different Components on the Mip-NeRF360 Dataset~\cite{mip360}, Which Contains Over $400K$ Anchors on Average. All Times Are Measured in Seconds, With Each Component's Percentage Contribution Indicated in Parentheses.}
    \vspace{-5pt}
    \renewcommand{\arraystretch}{1.30}  
    \setlength{\tabcolsep}{4.5pt}  
    \begin{tabular}{c|c|c|ccccccc}
        \toprule[2pt]
        $\lambda$   & \# \textbf{Valid Anchors}  & \textbf{Total Time} (s) & $\bm{x}^a$ & $\bm{f}^{a}$ & $\bm{l}$ & $\bm{o}$ & $\mathcal{H}$ & $\bm{m}$ & Others  \\ \toprule
        $0.5e-3$& 491852 & 18.10 & 3.27 (18\%) & 8.31 (46\%) & 2.82 (16\%) & 3.13 (17\%) & 0.01 (0\%) & 0.01 (0\%) & 0.55 (3\%) \\
        $1e-3$ & 449658 & 15.19 & 2.93 (19\%) & 6.95 (46\%) & 2.34 (15\%) & 2.84 (19\%) & 0.01 (0\%) & 0.01 (0\%) & 0.50 (3\%) \\
        $2e-3$ & 396485 & 12.17 & 2.57 (21\%) & 5.59 (46\%) & 1.86 (15\%) & 1.97 (16\%) & 0.01 (0\%) & 0.01 (0\%) & 0.33 (3\%) \\
        $3e-3$ & 359629 & 10.59 & 2.29 (22\%) & 4.82 (46\%) & 1.61 (15\%) & 1.56 (15\%) & 0.01 (0\%) & 0.01 (0\%) & 0.31 (3\%) \\
        $4e-3$ & 342049 & 9.80  & 2.20 (22\%) & 4.38 (45\%) & 1.46 (15\%) & 1.37 (14\%) & 0.01 (0\%) & 0.01 (0\%) & 0.26 (3\%) \\ \bottomrule[2pt]
    \end{tabular}
    \label{tab:encoding_time}
    \vspace{-5pt}
\end{table*}

\begin{table*}[t]
    \small
    \centering
    \caption{\textbf{Decoding Time} for Different Components on the Mip-NeRF360 Dataset~\cite{mip360}, Which Contains Over $400K$ Anchors on Average. All Times Are Measured in Seconds, With Each Component's Percentage Contribution Indicated in Parentheses.}
    \vspace{-5pt}
    \renewcommand{\arraystretch}{1.30}  
    \setlength{\tabcolsep}{4.5pt}  
    \begin{tabular}{c|c|c|ccccccc}
        \toprule[2pt]
        $\lambda$   & \# \textbf{Valid Anchors}  & \textbf{Total Time} (s) & $\bm{x}^a$ & $\bm{f}^{a}$ & $\bm{l}$ & $\bm{o}$ & $\mathcal{H}$ & $\bm{m}$ & Others  \\ \toprule
        $0.5e-3$& 491852 & 30.86 & 1.18 (4\%) & 14.97 (49\%) & 7.46 (24\%) & 6.55 (21\%) & 0.01 (0\%) & 0.02 (0\%) & 0.67 (2\%) \\
        $1e-3$ & 449658 & 25.62 & 1.09 (4\%) & 12.59 (49\%) & 6.35 (25\%) & 5.07 (20\%) & 0.01 (0\%) & 0.02 (0\%) & 0.51 (2\%) \\
        $2e-3$ & 396485 & 20.05 & 0.96 (5\%) & 10.06 (50\%) & 5.06 (25\%) & 3.55 (18\%) & 0.01 (0\%) & 0.02 (0\%) & 0.39 (2\%) \\
        $3e-3$ & 359629 & 17.22 & 0.88 (5\%) & 8.73 (51\%)  & 4.32 (25\%) & 2.91 (17\%) & 0.01 (0\%) & 0.01 (0\%) & 0.35 (2\%) \\
        $4e-3$ & 342049 & 15.77 & 0.83 (5\%) & 8.03 (51\%)  & 3.98 (25\%) & 2.61 (17\%) & 0.01 (0\%) & 0.02 (0\%) & 0.29 (2\%)  \\ \bottomrule[2pt]
    \end{tabular}
    \label{tab:decoding_time}
    \vspace{-5pt}
\end{table*}

\subsection{Mask Ratio Analysis}
\label{subsec:mask_ratio}

We present statistical data on the mask ratio of the adaptive offset masking in Tab.~\ref{tab:mask_ratio}. The valid ratio $r(\cdot)$ denotes the ratio of value $1$ in the mask, which indicates the corresponding anchor/Gaussian is valid. The 2nd and 3rd columns indicate that while the total number of anchors exceeds $560k$ due to the large scale of the dataset, only a subset of these anchors are valid. As $\lambda$ increases, stricter rate constraints lead to a decrease in the mask ratio. A similar trend is observed for Gaussians (offsets) in the 4th column. Since each anchor contains $K=10$ Gaussians, an anchor is considered valid if even just one of its Gaussians is valid. This results in the valid ratio of Gaussians ($r(\text{Gaussian})$) is significantly smaller than that of anchors ($r(\text{anchor})$).
Moreover, the value $\frac{r(\text{Gaussian})}{r(\text{anchor})}$ decreases as $\lambda$ increases, which indicates a reduced proportion of valid Gaussians within valid anchors, meaning growing positional redundancies.
Overall, leveraging this mask information, we effectively eliminate \textit{invalid anchors} and \textit{invalid offsets in valid anchors}, achieving a compact representation.

\begin{table}[t]
    \small
    \centering
    \caption{\textbf{Storage Size of Different Components} on the Mip-NeRF360 Dataset~\cite{mip360}. All Sizes Are Measured in \textbf{MB}.}
    \renewcommand{\arraystretch}{1.20}  
    \setlength{\tabcolsep}{3.5pt}  
    \begin{tabular}{c|c|ccccccc}
        \toprule[2pt]
        $\lambda$   & \textbf{Total Size}   & $\bm{x}^a$ & $\bm{f}^a$ & $\bm{l}$ & $\bm{o}$ & $\mathcal{H}$ & $\bm{m}$ & MLP \\ \toprule
        $0.5e-3$ & 18.48 & 0.91 & 9.54 & 2.52 & 4.49 & 0.12 & 0.56 & 0.33 \\
        $1e-3$ & 14.73 & 0.83 & 7.52 & 2.18 & 3.27 & 0.12 & 0.48 & 0.33 \\
        $2e-3$ & 11.18 & 0.75 & 5.61 & 1.81 & 2.19 & 0.11 & 0.38 & 0.33 \\
        $3e-3$ & 9.35  & 0.70 & 4.61 & 1.60 & 1.69 & 0.10 & 0.32 & 0.33 \\
        $4e-3$ & 8.34  & 0.67 & 4.00 & 1.48 & 1.47 & 0.09 & 0.30 & 0.33 \\ \bottomrule[2pt]
    \end{tabular}
    \label{tab:total_param_size}
    \vspace{-5pt}
\end{table}

\begin{table}[t]
    \normalsize
    \centering
    \caption{\textbf{Per-Parameter Bits of Different Components} on the Mip-NeRF360 Dataset~\cite{mip360}. All Sizes Are Measured in \textbf{Bits}.}
    \renewcommand{\arraystretch}{1.20}
    \setlength{\tabcolsep}{5pt}
    \begin{tabular}{c|ccccc}
        \toprule[2pt]
        $\lambda$   &  $\bm{x}^a$ & $\bm{f}^a$ & $\bm{l}$ & $\bm{o}$ & $\bm{m}$ \\ \toprule
        $0.5e-3$ & 5.19 & 3.23 & 7.15 & 6.72 & 0.94      \\
        $1e-3$ & 5.21 & 2.77 & 6.75 & 6.45 & 0.88      \\
        $2e-3$ & 5.34 & 2.32 & 6.35 & 6.20 & 0.80      \\
        $3e-3$ & 5.48 & 2.09 & 6.19 & 6.08 & 0.74      \\
        $4e-3$ & 5.53 & 1.90 & 6.00 & 5.93 & 0.72     \\ \bottomrule[2pt]
    \end{tabular}
    \label{tab:per_param_size}
    \vspace{-5pt}
\end{table}

\vspace{-10pt}
\begin{table}[t]
    \normalsize
    \centering
    \caption{Ratio of Valid Anchors and Gaussians (\ie, Offsets) on the Mip-NeRF360 Dataset~\cite{mip360}. Each Anchor Has $K=10$ Gaussians.}
    \renewcommand{\arraystretch}{1.00}
        \begin{tabular}{c|ccc}
        \toprule[2pt]
         $\lambda$ & \# \textbf{Total Anchors} & $r(\text{anchor})$ & $r(\text{Gaussian})$   \\ \toprule
        $0.5e-3$  & 560425  & 0.865  & 0.314 \\
        $1e-3$  & 561429  & 0.790  & 0.240 \\
        $2e-3$  & 564852  & 0.695  & 0.168 \\
        $3e-3$  & 570567  & 0.628  & 0.132 \\
        $4e-3$ & 571923  & 0.596  & 0.118 \\ \bottomrule[2pt]
        \end{tabular}
    \label{tab:mask_ratio}
    \vspace{-10pt}
\end{table}

\begin{table}[t]
    \centering
    \small
    \caption{\textbf{Encoding and Decoding Time} across all Datasets. ``V. A.'' Denotes ``Averaged Number of Valid Anchors''.
    }
    \setlength{\tabcolsep}{2pt}
    \renewcommand{\arraystretch}{1.1}
    \begin{tabular}{c|c|cc|cc}
    \toprule[2pt]
    \multirow{2}{*}{\textbf{Datasets}} & \multirow{2}{*}{\textbf{V. A.}} & \multicolumn{2}{c|}{\textbf{Enc. Time} (s)} & \multicolumn{2}{c}{\textbf{Dec. Time} (s)} \\ \cline{3-6} 
                          &   & {\small lowrate}        & {\small highrate}       & {\small lowrate}    & {\small highrate}  \\ \toprule[1pt]
    Synthetic-NeRF~\cite{NeRF}          &       38467      & 0.80              & 1.57              &      1.18     &      2.36     \\
    Mip-NeRF360~\cite{mip360}           &     407934       & 9.80              & 18.10              &      15.77     &      30.86     \\
    Tank\&Temples~\cite{tant}           &      240915      & 6.01              & 9.01              &      9.58     &      14.20     \\
    DeepBlending~\cite{deepblending}      &      145773           & 3.35              & 5.17              &      4.92     &      7.86     \\
    BungeeNeRF~\cite{BungeeNeRF}         &      460713      & 12.44              & 18.62              &      18.67     &      28.58     \\
    \bottomrule[2pt]
    \end{tabular}
    \label{tab:enc_dec_time_all}
    \vspace{-10pt}
\end{table}

\subsection{Decomposition of Storage Size of Difference Components}
\label{subsec:decom_size}

This subsection provides a detailed analysis of the storage size associated with various attribute components, from both a macroscopic and microscopic perspective. It is important to note that only valid anchors require encoding, while invalid anchors are directly discarded. Specifically, for offsets $\bm{o}$, only the valid offsets are encoded for calculation, which is different from other attributes.

\begin{itemize}
    \item From a macroscopic perspective, the total storage size for each component is summarized in Tab.~\ref{tab:total_param_size}. As $\lambda$ increases, the rate constraint is stricter, leading to an overall reduction in storage size. Notably, the storage size of the offset $\bm{o}$ decreases most significantly ($3\times$ reduction), due to two factors: a reduction in per-parameter bits (see Tab.~\ref{tab:per_param_size}) and the decreased ratio of valid offsets within valid anchors (indicated by $\frac{r(\text{Gaussian})}{r(\text{anchor})}$ in Tab.~\ref{tab:mask_ratio}), leading to fewer parameters.
    \item From a microscopic perspective, the per-parameter bits for each component are in Tab.~\ref{tab:per_param_size}. While most attributes show a decreasing trend as $\lambda$ increases, the anchor location $\bm{x}^a$ behaves differently. Since $\bm{x}^a$ is compressed using GPCC, more anchor locations in high-rate segments lead to greater positional redundancies, lowering the per-parameter bits. For the mask $\bm{m}$, as $\lambda$ increases, the ratio of valid offsets within valid anchors decreases, making zeros more dominant in the offset mask and consequently reducing entropy for each parameter.
\end{itemize}

\subsection{Decomposition of Coding Time of Difference Components}
\label{subsec:decom_coding}

In this subsection, we analyze the coding time of different attributes in HAC++. Tab.~\ref{tab:encoding_time} and~\ref{tab:decoding_time} present the times for encoding and decoding, respectively. Despite the large scale of the Mip-NeRF360 dataset, with over $400K$ valid anchors on average, both encoding and decoding processes remain efficient within 30 seconds.
The anchor locations $\bm{x}^a$ are compressed using GPCC, leading to a relatively long encoding time due to the complex RD search process, but a significantly shorter decoding time. Conversely, the other attributes are encoded and decoded via AE, where the decoding phase involves index searching, increasing its complexity.
The ``Others'' category accounts for the time taken for model-related operations, such as MLP processing. Additionally, Table~\ref{tab:enc_dec_time_all} reports the coding times across all datasets. For simpler datasets, the coding time is much shorter due to the smaller number of anchors.
Overall, the coding process is efficient and could be further optimized through advanced codec techniques, which we consider an engineering task for future work.

\subsection{Training and Rendering Efficiency}
\label{subsec:train_render_efficiency}

\begin{figure*}[t]
    \centering
    \includegraphics[width=0.93\linewidth]{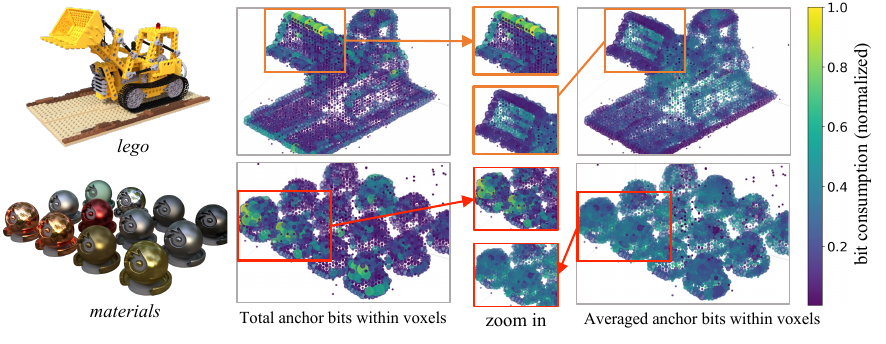}
    \caption{Visualization of bit allocation across anchors for the scenes ``lego'' and ``materials'' on the Synthetic-NeRF dataset~\cite{NeRF}. The 3D space is voxelized, with each voxel represented by a ball and the radius of a ball indicating the number of anchors in the voxel. 
    For the 2nd column, 
    the color of a ball indicates the \textit{total} bit consumption of all anchors in the voxel, while for the 4th column, 
    the color represents the \textit{averaged} bit consumption per anchor within a voxel. The 3rd column gives zoom-in views. It shows more anchors are allocated to important regions while the bit consumption for each anchor is smooth.}
    \label{fig:bit_allocation}
\end{figure*}

\begin{table}[t]
    \centering
    \small
    \caption{\textbf{Training Time and Peak GPU Memory Usage} of Our Approach Compared to Previous Methods on the Mip-NeRF360 Dataset~\cite{mip360}. 
    }
    \begin{tabular}{c|cc|cc}
    \toprule[2pt]
    \multirow{2}{*}{\textbf{Methods}} & \multicolumn{2}{c|}{\textbf{Training Time} (s)} & \multicolumn{2}{c}{\textbf{Peak GPU Mem} (GB)} \\ \cline{2-5} 
                             & {\small lowrate}        & {\small highrate}       & {\small lowrate}    & {\small highrate}  \\ \toprule[1pt]
    3DGS~\cite{3DGS}                     & \multicolumn{2}{c|}{1590}          & \multicolumn{2}{c}{12.00}   \\
    Scaffold-GS~\cite{scaffold}                 & \multicolumn{2}{c|}{1286}          & \multicolumn{2}{c}{9.69}    \\ \hline
    \textbf{HAC++}                    & 2384        & 2278         &    10.87    &   11.66     \\
    \bottomrule[2pt]
    \end{tabular}
    \label{tab:time_train}
\end{table}

\begin{table}[t]
    \centering
    \small
    \caption{\textbf{Gaussian Count and FPS} of Our Approach Compared to Previous Methods on the Mip-NeRF360 Dataset~\cite{mip360}. 
    }
    \begin{tabular}{c|cc|cc}
    \toprule[2pt]
    \multirow{2}{*}{\textbf{Methods}} & \multicolumn{2}{c|}{\# \textbf{Valid Gaussians}} & \multicolumn{2}{c}{\textbf{FPS}} \\ \cline{2-5} 
                             & {\small lowrate}    & {\small highrate} & {\small lowrate}    & {\small highrate}  \\ \toprule[1pt]
    3DGS~\cite{3DGS}                     & \multicolumn{2}{c|}{3175k}   & \multicolumn{2}{c}{99} \\
    Scaffold-GS~\cite{scaffold}                 & \multicolumn{2}{c|}{5674k}    & \multicolumn{2}{c}{135} \\ \hline
    \textbf{HAC++}                    & 682k       & 1853k       & 151       & 130    \\
    \bottomrule[2pt]
    \end{tabular}
    \label{tab:time_test}
\end{table}

In this subsection, we evaluate the training and rendering efficiency of our HAC++ method, as shown in Tab.~\ref{tab:time_train} and~\ref{tab:time_test}. For the two base methods, 3DGS~\cite{3DGS} and Scaffold-GS~\cite{scaffold}, Scaffold-GS demonstrates lower training times and faster rendering FPS, despite having a higher Gaussian count. This performance advantage is attributed to its pre-filtering scheme, which skips computations for out-of-view anchors (and Gaussians) and precomputes colors to avoid the complex SH calculations. These features enable Scaffold-GS to achieve faster rendering speeds compared to 3DGS. Since HAC++ is built upon Scaffold-GS, it inherits these efficient features.
\textbf{During training}, as shown in Tab.~\ref{tab:time_train}, the inclusion of context models in HAC++ results in an $81\%$ increase in training time compared to Scaffold-GS. However, HAC++ still maintains a fast training speed. Notably, the training time remains consistent across different rates because, during training, masks are applied to all Gaussians/anchors to preserve gradients, regardless of their validity. Consequently, forward and backward passes for both valid and invalid Gaussians/anchors are executed, making the time consumption consistent. Additionally, due to the sampling strategy employed, the peak GPU memory usage remains reasonable, with only a $16\%$ average increase over Scaffold-GS.
\textbf{During inference}, as shown in Tab.~\ref{tab:time_test}, HAC++ benefits from its context modeling design, enabling the removal of the hash grid after decoding $\mathcal{A}$. This design eliminates the need for additional operations during rendering. Furthermore, HAC++ exhibits FPS improvements over Scaffold-GS, particularly at low rates. This improvement arises from the adaptive masking design, which prunes invalid Gaussians/anchors, thereby accelerating the rendering process.

\subsection{Visualization of Bit Allocation}
\label{subsec:visual_bit}
While HAC++ measures the parameters' bit consumption, we are interested in the bit allocation across different local areas in the space. 
In Fig.~\ref{fig:bit_allocation}, we utilize scenes in Synthetic-NeRF dataset~\cite{NeRF} for visualization, and represent bit allocation conditions by voxelized colored balls. As observed from the 2nd column of visualized sub-figures, the model tends to allocate more total bits to areas with complex appearances or sharp edges. For instance, edge areas in ``lego'' and specular objects in ``materials'' exhibit higher total bit consumption due to the complex textures. The analysis of the 4th column from an averaging viewpoint reveals varied trends in bit consumption per anchor. In high bit-consumption voxels, creating more anchors for precise modeling averages the bit per anchor, smoothing or reducing bit consumption for each. This aligns with our assumption that anchors demonstrate inherent consistency in the 3D space where nearby anchors exhibit similar values of attributes, making it easier for the hash grid to accurately estimate their value probabilities.

\begin{figure}[t]
    \centering
    \includegraphics[width=0.90\linewidth]{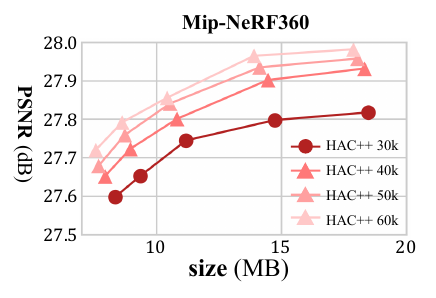}
    \vspace{-10pt}
    \caption{Training with different number of iterations on the Mip-NeRF360 dataset~\cite{mip360}.}
    \label{fig:diff_iterations}
\end{figure}

\begin{table}[t]
    \centering
    \caption{Training Time and BD-Rate Changes Across Different Iterations. Experiments Are Conducted on the Mip-NeRF360 Dataset~\cite{mip360}. Negative BD-Rate Values Indicate Decreases in Relative Size Compared to the Anchor Method (HAC++) at the Same Fidelity, Representing Performance Improvements.}
    \begin{tabular}{l|c|c}
        \toprule[2pt]
        \textbf{Training Iterations}                  & \textbf{Training Time} (s)  &  \textbf{BD-rate} $\downarrow$ \\ \toprule
        HAC++ 60k   & 5294 &  $-32.9\%$     \\
        HAC++ 50k   & 4224 &  $-25.6\%$     \\
        HAC++ 40k   & 3307 &  $-17.5\%$     \\ \hline
        \textbf{HAC++ 30k} (anchor method)         & 2292 &   $0.0\%$    \\   \bottomrule[2pt]
    \end{tabular}
    \label{tab:diff_iterations}
\end{table}

\subsection{Training with Different Number of Iterations}
\label{subsec:training_diff_iter}

In this subsection, we analyze the impact of varying number of training iterations on performance, as illustrated in Fig.~\ref{fig:diff_iterations} and Tab.~\ref{tab:diff_iterations}. Leveraging a per-scene optimization scheme, HAC++ exhibits enhanced performance with increased training iterations, but at the expense of extended training time. Notably, a substantial BD-rate improvement of $-32.9\%$ is observed when comparing 60k iterations to the default 30k iterations.
When comparing performance under the same $\lambda$ value across different training curves, the primary gains are attributed to fidelity improvements, with minimal reductions in storage size. However, as the total number of iterations increases, the incremental benefit of additional iterations becomes diminished.

Overall, selecting appropriate training iterations is crucial for achieving a desirable trade-off between training time and compression performance, which can be tailored to specific applications or computational resources.

\section{Supplementary: Per-Scene Result of HAC++}
\label{sec:per_scene}
We present per-scene results of our HAC++ method across multiple fidelity metrics (\ie, PSNR, SSIM~\cite{ssim} and LPIPS~\cite{LPIPS}) over all datasets. Results are presented in the Appendix.

\section{Conclusion and limitation}

We have explored the relationship between unorganized, sparse Gaussians (or anchors) and well-structured hash grids, leveraging their mutual information to achieve compressed 3DGS representations. Through the integration of a Hash-grid Assisted Context (HAC) module and an intra-anchor context model, our HAC++ method achieves the SoTA compression performance. Extensive experiments validate the effectiveness of HAC++ and its technical components through comprehensive analyses. By addressing the significant challenge of large storage requirements in 3DGS representations, our work paves the way for their deployment in large-scale scenes.

\noindent\textbf{Limitation.}
The main limitation of HAC++ lies in its increased training time compared to the base method, Scaffold-GS, due to the additional loss term and the incorporation of context models. Future work could explore lightweight context model designs to alleviate this issue. Furthermore, HAC++ establishes relationships among anchors indirectly through an intermediate hash grid. Investigating approaches that directly model relationships among anchors could provide an alternative strategy for redundancy elimination.

\section*{Acknowledgement}
The paper is supported in part by The National Natural Science Foundation of China (No. 62325109, U21B2013). 

\noindent MH is supported by funding from The Australian Research Council Discovery Program DP230101176.

% \balance
\bibliographystyle{IEEEtran}
\bibliography{main}

\begin{IEEEbiographynophoto}{Yihang Chen}
received the BS degree from Electronic Information Engineering, Dalian University of Technology, in 2021. He is currently pursuing the Ph.D. degree with the joint program of Shanghai Jiao Tong University and Monash University. His current research interests include 3D vision and data compression techniques.
\end{IEEEbiographynophoto}

\begin{IEEEbiographynophoto}{Qianyi Wu}
received the BSc and MSc degrees from the Special Class for the Gifted Youth and School of Mathematical Sciences, University of Science and Technology of China in 2016 and 2019, respectively.  He is currently pursuing the Ph.D. degree with the Faculty of Information Technology, the Monash University. His current research interest includes computer vision and computer graphics.
\end{IEEEbiographynophoto}

\begin{IEEEbiographynophoto}{Weiyao Lin}
(Senior Member, IEEE) received the B.E. and M.E. degrees from Shanghai Jiao Tong University, Shanghai, China, in 2003 and 2005, respectively, and the Ph.D. degree from the University of Washington, Seattle, WA, USA, in 2010, all in electrical engineering. He is currently a Professor with the Department of Electronic Engineering, Shanghai Jiao Tong University. He has authored or coauthored more than 100 technical articles on top journals/conferences including the IEEE TRANSACTIONS ON PATTERN ANALYSIS AND MACHINE INTELLIGENCE, International Journal of Computer Vision, the IEEE TRANSACTIONS ON IMAGE PROCESSING, CVPR, and ICCV. He holds more than 20 patents. His research interests include video/image analysis, computer vision, and video/image processing applications .

\end{IEEEbiographynophoto}

\begin{IEEEbiographynophoto}{Mehrtash Harandi}
is an associate professor with the Department of Electrical and Computer Systems Engineering at Monash University. He is also a contributing research scientist in the Machine Learning Research Group (MLRG) at Data61/CSIRO. His current research interests include theoretical and computational methods in machine learning, computer vision, and Riemannian geometry.
\end{IEEEbiographynophoto}

\begin{IEEEbiographynophoto}{Jianfei Cai}
(S'98-M'02-SM'07-F’21) received his PhD degree from the University of Missouri-Columbia. He is currently a Professor at Faculty of IT, Monash University, where he had served as the inaugural Head for the Data Science \& AI Department. Before that, he was Head of Visual and Interactive Computing Division and Head of Computer Communications Division in Nanyang Technological University (NTU). His major research interests include computer vision, deep learning and multimedia. He is a co-recipient of paper awards in ACCV, ICCM, IEEE ICIP and MMSP, and the winner of Monash FIT’s Dean’s Researcher of the Year Award. He is currently on the editorial board of TPAMI and IJCV. He has served as an Associate Editor for IEEE T-IP, T-MM, and T-CSVT as well as serving as Area Chair for CVPR, ICCV, ECCV, ACM Multimedia, IJCAI, ICME, ICIP, and ISCAS. He was the Chair of IEEE CAS VSPC-TC during 2016-2018. He had served as the leading TPC Chair for IEEE ICME 2012 and the best paper award committee chair \& co-chair for IEEE T-MM 2020 \& 2019. He is the leading General Chair for ACM Multimedia 2024, and a Fellow of IEEE.
\end{IEEEbiographynophoto}

\clearpage

\setcounter{page}{1}
\twocolumn[
\begin{center}
  \textbf{{\Large --Appendix--}}
\end{center}
\vspace{15pt}
]
\vspace{15pt}

\renewcommand\thesection{\Alph{section}}
\renewcommand\thetable{\Alph{table}}
\renewcommand\thefigure{\Alph{figure}}
\setcounter{section}{0}
\setcounter{table}{0}
\setcounter{figure}{0}

% In this Appendix, we present per-scene results of our HAC++ method across multiple fidelity metrics (\ie, PSNR, SSIM~\cite{ssim} and LPIPS~\cite{LPIPS}) over all datasets, as shown from Tab.~\ref{tab:per_scene_blender} to Tab.~\ref{tab:per_scene_bungee}.

\begin{table}[ht]
\caption{HAC++'s Results on the Synthetic-NeRF Dataset~\cite{NeRF} for Different $\lambda$ Values.}
    \centering
    \setlength{\tabcolsep}{4pt}  % 列间距
    \begin{tabular}{c|c|ccc|c}
     \toprule[2pt]
       $\lambda$ & Scenes  & PSNR $\uparrow$ & SSIM $\uparrow$ & LPIPS $\downarrow$ & Size (MB) $\downarrow$ \\ \toprule
    \multirow{8}{*}{$4e-3$} & chair & 33.90 & 0.980 & 0.021 & 0.65 \\
     & drums & 26.17 & 0.950 & 0.045 & 0.90 \\
     & ficus & 34.53 & 0.984 & 0.016 & 0.64 \\
     & hotdog & 36.53 & 0.979 & 0.035 & 0.45 \\
     & lego & 34.33 & 0.976 & 0.027 & 0.80 \\
     & materials & 30.22 & 0.959 & 0.046 & 0.87 \\
     & mic & 35.21 & 0.989 & 0.012 & 0.44 \\
     & ship & 31.26 & 0.903 & 0.125 & 1.30 \\ \midrule
    \multirow{8}{*}{$3e-3$} & chair & 34.40 & 0.982 & 0.018 & 0.76 \\
     & drums & 26.27 & 0.951 & 0.043 & 1.20 \\
     & ficus & 34.74 & 0.984 & 0.015 & 0.70 \\
     & hotdog & 36.87 & 0.980 & 0.032 & 0.54 \\
     & lego & 34.64 & 0.977 & 0.025 & 0.97 \\
     & materials & 30.35 & 0.960 & 0.044 & 0.97 \\
     & mic & 35.64 & 0.990 & 0.010 & 0.50 \\
     & ship & 31.34 & 0.903 & 0.124 & 1.42 \\ \midrule
    \multirow{8}{*}{$2e-3$} & chair & 34.79 & 0.984 & 0.016 & 0.94 \\
     & drums & 26.25 & 0.951 & 0.043 & 1.25 \\
     & ficus & 34.80 & 0.985 & 0.014 & 0.84 \\
     & hotdog & 37.27 & 0.982 & 0.028 & 0.60 \\
     & lego & 35.10 & 0.979 & 0.022 & 1.08 \\
     & materials & 30.49 & 0.961 & 0.041 & 1.15 \\
     & mic & 36.05 & 0.991 & 0.009 & 0.61 \\
     & ship & 31.43 & 0.903 & 0.119 & 1.88 \\ \midrule
    \multirow{8}{*}{$1e-3$} & chair & 35.34 & 0.986 & 0.014 & 1.21 \\
     & drums & 26.41 & 0.952 & 0.041 & 1.79 \\
     & ficus & 35.21 & 0.986 & 0.013 & 1.11 \\
     & hotdog & 37.69 & 0.983 & 0.025 & 0.78 \\
     & lego & 35.52 & 0.981 & 0.019 & 1.49 \\
     & materials & 30.64 & 0.962 & 0.040 & 1.48 \\
     & mic & 36.51 & 0.991 & 0.008 & 0.79 \\
     & ship & 31.52 & 0.905 & 0.114 & 2.55 \\ \midrule
    \multirow{8}{*}{$0.5e-3$} & chair & 35.60 & 0.986 & 0.012 & 1.64 \\
     & drums & 26.48 & 0.952 & 0.041 & 2.47 \\
     & ficus & 35.25 & 0.986 & 0.013 & 1.42 \\
     & hotdog & 37.89 & 0.984 & 0.023 & 0.98 \\
     & lego & 35.77 & 0.982 & 0.018 & 1.83 \\
     & materials & 30.71 & 0.962 & 0.038 & 1.90 \\
     & mic & 36.79 & 0.992 & 0.008 & 1.09 \\
     & ship & 31.54 & 0.904 & 0.111 & 3.43 \\ \toprule
     \bottomrule
    $4e-3$ & \textbf{AVG} & 32.77 & 0.965 & 0.041 & 0.76 \\
    $3e-3$ & \textbf{AVG} & 33.03 & 0.966 & 0.039 & 0.88 \\
    $2e-3$ & \textbf{AVG} & 33.27 & 0.967 & 0.037 & 1.04 \\
    $1e-3$ & \textbf{AVG} & 33.60 & 0.968 & 0.034 & 1.40 \\
    $0.5e-3$ & \textbf{AVG} & 33.76 & 0.969 & 0.033 & 1.84 \\
     \toprule[2pt]
    \end{tabular}
    \label{tab:per_scene_blender}
\end{table}

\begin{table}[ht]
\caption{HAC++'s Results on the DeepBlending Dataset~\cite{deepblending} for Different $\lambda$ Values.}
    \centering
    \setlength{\tabcolsep}{4pt}  % 列间距
    \begin{tabular}{c|c|ccc|c}
     \toprule[2pt]
       $\lambda$ & Scenes  & PSNR $\uparrow$ & SSIM $\uparrow$ & LPIPS $\downarrow$ & Size (MB) $\downarrow$ \\ \toprule
    \multirow{2}{*}{$4e-3$} & playroom & 30.69 & 0.909 & 0.268 & 2.42 \\
     & drjohnson & 29.63 & 0.905 & 0.264 & 3.40 \\ \midrule
    \multirow{2}{*}{$3e-3$} & playroom & 30.82 & 0.911 & 0.263 & 2.78 \\
     & drjohnson & 29.75 & 0.906 & 0.260 & 3.91 \\ \midrule
    \multirow{2}{*}{$2e-3$} & playroom & 30.79 & 0.911 & 0.262 & 3.29 \\
     & drjohnson & 29.73 & 0.907 & 0.257 & 4.61 \\ \midrule
    \multirow{2}{*}{$1e-3$} & playroom & 30.93 & 0.913 & 0.255 & 4.35 \\
     & drjohnson & 29.76 & 0.908 & 0.253 & 6.21 \\ \toprule
     \bottomrule
    $4e-3$ & \textbf{AVG} & 30.16 & 0.907 & 0.266 & 2.91 \\
    $3e-3$ & \textbf{AVG} & 30.28 & 0.909 & 0.262 & 3.34 \\
    $2e-3$ & \textbf{AVG} & 30.26 & 0.909 & 0.260 & 3.95 \\
    $1e-3$ & \textbf{AVG} & 30.34 & 0.911 & 0.254 & 5.28 \\
     \toprule[2pt]
    \end{tabular}
    \label{tab:per_scene_blending}
\end{table}

\begin{table}[t]
\caption{HAC++'s Results on the Tank\&Temples Dataset~\cite{tant} for Different $\lambda$ Values.}
    \centering
    \setlength{\tabcolsep}{4pt}  % 列间距
    \begin{tabular}{c|c|ccc|c}
     \toprule[2pt]
       $\lambda$ & Scenes  & PSNR $\uparrow$ & SSIM $\uparrow$ & LPIPS $\downarrow$ & Size (MB) $\downarrow$ \\ \toprule
    \multirow{2}{*}{$4e-3$} & truck & 25.89 & 0.883 & 0.156 & 5.67 \\
     & train & 22.54 & 0.815 & 0.225 & 4.69 \\ \midrule
    \multirow{2}{*}{$3e-3$} & truck & 25.98 & 0.884 & 0.153 & 6.53 \\
     & train & 22.57 & 0.818 & 0.219 & 5.25 \\ \midrule
    \multirow{2}{*}{$2e-3$} & truck & 26.06 & 0.886 & 0.148 & 7.64 \\
     & train & 22.60 & 0.821 & 0.213 & 6.21 \\ \midrule
    \multirow{2}{*}{$1e-3$} & truck & 26.05 & 0.887 & 0.147 & 9.71 \\
     & train & 22.58 & 0.821 & 0.210 & 7.55 \\ \midrule
     \bottomrule
    $4e-3$ & \textbf{AVG} & 24.22 & 0.849 & 0.190 & 5.18 \\
    $3e-3$ & \textbf{AVG} & 24.28 & 0.851 & 0.186 & 5.89 \\
    $2e-3$ & \textbf{AVG} & 24.33 & 0.853 & 0.181 & 6.92 \\
    $1e-3$ & \textbf{AVG} & 24.32 & 0.854 & 0.178 & 8.63 \\
     \toprule[2pt]
    \end{tabular}
    \label{tab:per_scene_tandt}
\end{table}

\begin{table}[ht]
\caption{HAC++'s Results on the Mip-NeRF360 Dataset~\cite{mip360} for Different $\lambda$ Values.}
    \centering
    \setlength{\tabcolsep}{4pt}  % 列间距
    \begin{tabular}{c|c|ccc|c}
     \toprule[2pt]
       $\lambda$ & Scenes  & PSNR $\uparrow$ & SSIM $\uparrow$ & LPIPS $\downarrow$ & Size (MB) $\downarrow$ \\ \toprule
    \multirow{9}{*}{$4e-3$} & bicycle & 25.08 & 0.732 & 0.289 & 12.84 \\
     & garden & 27.19 & 0.832 & 0.179 & 12.83 \\
     & stump & 26.56 & 0.757 & 0.289 & 9.40 \\
     & room & 31.75 & 0.920 & 0.217 & 3.41 \\
     & counter & 29.37 & 0.910 & 0.203 & 4.69 \\
     & kitchen & 31.20 & 0.923 & 0.137 & 5.26 \\
     & bonsai & 32.66 & 0.944 & 0.190 & 5.35 \\
     & flower & 21.26 & 0.569 & 0.393 & 10.68 \\
     & treehill & 23.31 & 0.638 & 0.379 & 10.57 \\ \midrule
    \multirow{9}{*}{$3e-3$} & bicycle & 25.09 & 0.734 & 0.285 & 14.36 \\
     & garden & 27.21 & 0.835 & 0.172 & 14.37 \\
     & stump & 26.61 & 0.758 & 0.283 & 10.47 \\
     & room & 31.83 & 0.921 & 0.213 & 3.84 \\
     & counter & 29.48 & 0.912 & 0.199 & 5.28 \\
     & kitchen & 31.34 & 0.924 & 0.134 & 6.01 \\
     & bonsai & 32.73 & 0.946 & 0.188 & 5.86 \\
     & flower & 21.28 & 0.570 & 0.390 & 11.99 \\
     & treehill & 23.32 & 0.640 & 0.375 & 11.96 \\ \midrule
    \multirow{9}{*}{$2e-3$} & bicycle & 25.22 & 0.738 & 0.276 & 17.24 \\
     & garden & 27.36 & 0.842 & 0.159 & 17.22 \\
     & stump & 26.62 & 0.761 & 0.277 & 12.29 \\
     & room & 32.00 & 0.924 & 0.206 & 4.54 \\
     & counter & 29.64 & 0.915 & 0.193 & 6.26 \\
     & kitchen & 31.51 & 0.927 & 0.129 & 7.14 \\
     & bonsai & 32.88 & 0.947 & 0.185 & 6.83 \\
     & flower & 21.22 & 0.571 & 0.385 & 14.42 \\
     & treehill & 23.26 & 0.642 & 0.366 & 14.65 \\ \midrule
    \multirow{9}{*}{$1e-3$} & bicycle & 25.04 & 0.740 & 0.268 & 23.09 \\
     & garden & 27.44 & 0.847 & 0.147 & 22.14 \\
     & stump & 26.60 & 0.762 & 0.269 & 16.47 \\
     & room & 32.04 & 0.927 & 0.199 & 5.99 \\
     & counter & 29.71 & 0.917 & 0.188 & 8.06 \\
     & kitchen & 31.70 & 0.930 & 0.125 & 9.23 \\
     & bonsai & 33.06 & 0.949 & 0.181 & 8.84 \\
     & flower & 21.31 & 0.575 & 0.378 & 19.03 \\
     & treehill & 23.30 & 0.646 & 0.352 & 19.72 \\ \midrule
    \multirow{9}{*}{$0.5e-3$} & bicycle & 25.05 & 0.738 & 0.268 & 29.25 \\
     & garden & 27.49 & 0.850 & 0.140 & 27.22 \\
     & stump & 26.52 & 0.761 & 0.265 & 21.02 \\
     & room & 32.05 & 0.928 & 0.196 & 7.55 \\
     & counter & 29.81 & 0.919 & 0.183 & 9.73 \\
     & kitchen & 31.81 & 0.931 & 0.122 & 11.37 \\
     & bonsai & 33.17 & 0.950 & 0.179 & 11.15 \\
     & flower & 21.28 & 0.576 & 0.375 & 24.19 \\
     & treehill & 23.20 & 0.645 & 0.349 & 24.85 \\ \toprule
     \bottomrule
    $4e-3$ & \textbf{AVG} & 27.60 & 0.803 & 0.253 & 8.34 \\
    $3e-3$ & \textbf{AVG} & 27.65 & 0.805 & 0.249 & 9.35 \\
    $2e-3$ & \textbf{AVG} & 27.74 & 0.808 & 0.242 & 11.18 \\
    $1e-3$ & \textbf{AVG} & 27.80 & 0.810 & 0.234 & 14.73 \\
    $0.5e-3$ & \textbf{AVG} & 27.82 & 0.811 & 0.231 & 18.48 \\
     \toprule[2pt]
    \end{tabular}
    \label{tab:per_scene_mip}
\end{table}

\begin{table}[t]
\caption{HAC++'s Results on the BungeeNeRF Dataset~\cite{BungeeNeRF} for Different $\lambda$ Values.}
    \centering
    \setlength{\tabcolsep}{4pt}  % 列间距
    \begin{tabular}{c|c|ccc|c}
     \toprule[2pt]
       $\lambda$ & Scenes  & PSNR $\uparrow$ & SSIM $\uparrow$ & LPIPS $\downarrow$ & Size (MB) $\downarrow$ \\ \toprule
    \multirow{6}{*}{$4e-3$} & amsterdam & 26.85 & 0.871 & 0.220 & 14.05 \\
     & bilbao & 27.92 & 0.881 & 0.206 & 11.02 \\
     & hollywood & 24.30 & 0.753 & 0.348 & 11.22 \\
     & pompidou & 25.35 & 0.846 & 0.244 & 12.49 \\
     & quebec & 30.00 & 0.930 & 0.172 & 9.84 \\
     & rome & 26.25 & 0.865 & 0.222 & 11.91 \\ \midrule
    \multirow{6}{*}{$3e-3$} & amsterdam & 27.03 & 0.879 & 0.206 & 16.01 \\
     & bilbao & 28.09 & 0.887 & 0.191 & 12.59 \\
     & hollywood & 24.51 & 0.766 & 0.334 & 12.76 \\
     & pompidou & 25.55 & 0.852 & 0.235 & 14.81 \\
     & quebec & 30.19 & 0.934 & 0.164 & 11.39 \\
     & rome & 26.39 & 0.872 & 0.210 & 13.61 \\ \midrule
    \multirow{6}{*}{$2e-3$} & amsterdam & 27.22 & 0.888 & 0.189 & 19.35 \\
     & bilbao & 28.12 & 0.893 & 0.179 & 15.14 \\
     & hollywood & 24.71 & 0.784 & 0.312 & 15.37 \\
     & pompidou & 25.70 & 0.858 & 0.227 & 17.39 \\
     & quebec & 30.35 & 0.937 & 0.155 & 13.47 \\
     & rome & 26.67 & 0.881 & 0.199 & 16.07 \\ \midrule
    \multirow{6}{*}{$1e-3$} & amsterdam & 27.29 & 0.896 & 0.169 & 25.20 \\
     & bilbao & 27.89 & 0.894 & 0.169 & 19.58 \\
     & hollywood & 24.84 & 0.797 & 0.290 & 19.48 \\
     & pompidou & 25.73 & 0.860 & 0.219 & 22.57 \\
     & quebec & 30.51 & 0.941 & 0.146 & 17.49 \\
     & rome & 26.78 & 0.887 & 0.184 & 20.58 \\ \toprule
     \bottomrule
    $4e-3$ & \textbf{AVG} & 26.78 & 0.858 & 0.235 & 11.75 \\
    $3e-3$ & \textbf{AVG} & 26.96 & 0.865 & 0.223 & 13.53 \\
    $2e-3$ & \textbf{AVG} & 27.13 & 0.873 & 0.210 & 16.13 \\
    $1e-3$ & \textbf{AVG} & 27.17 & 0.879 & 0.196 & 20.82 \\
     \toprule[2pt]
    \end{tabular}
    \label{tab:per_scene_bungee}
\end{table}

\vfill

\end{document}